\documentclass[journal]{IEEEtran}

\usepackage{amsmath,amssymb,amsfonts}
\usepackage{graphicx}
\usepackage{textcomp}
\usepackage{cite}
\usepackage{times}
\usepackage{multirow,booktabs,enumitem,makecell,ragged2e,adjustbox,comment}
\usepackage[caption=true,font=footnotesize,labelfont=sf,textfont=rm]{subfig}
\usepackage{bm}
\usepackage[hidelinks]{hyperref}
\usepackage{cuted}    
\usepackage{capt-of}  

\usepackage{fancyhdr}

\makeatletter
\let\ps@IEEEtitlepagestyle\ps@fancy
\makeatother

\begin{document}

\title{From Fields to Splats: A Cross-Domain Survey of Real-Time Neural Scene Representations}

\author{%
Javed~Ahmad\IEEEauthorrefmark{1},
Penggang~Gao\IEEEauthorrefmark{1},
Donatien~Delehelle\IEEEauthorrefmark{1},
Mennuti~Canio\IEEEauthorrefmark{2},
Nikhil~Deshpande\IEEEauthorrefmark{3},
Jes\'us~Ortiz\IEEEauthorrefmark{1},
Darwin~G.~Caldwell\IEEEauthorrefmark{1},
Yonas~Teodros~Tefera\IEEEauthorrefmark{1}%
\thanks{\IEEEauthorrefmark{1}\;Advanced Robotics, Istituto Italiano di Tecnologia (IIT), Genova, GE 16163, Italy.}%
\thanks{\IEEEauthorrefmark{2}\;Istituto Nazionale per l'Assicurazione contro gli Infortuni sul Lavoro (INAIL), Roma, Italy.}%
\thanks{\IEEEauthorrefmark{3}\;School of Computer Science, University of Nottingham, UK.}%
\thanks{Corresponding authors: javed.ahmad@iit.it, yonas.tefera@iit.it}%
}

\IEEEtitleabstractindextext{%
\begin{abstract}
Neural scene representations such as Neural Radiance Fields (NeRF) and 3D Gaussian Splatting (3DGS) have transformed how 3D environments are modeled, rendered, and interpreted. NeRF introduced view-consistent photorealism via volumetric rendering; 3DGS has rapidly emerged as an explicit, efficient alternative that supports high-quality rendering, faster optimization, and integration into hybrid pipelines for enhanced photorealism and task-driven scene understanding.
This survey examines how 3DGS is being adopted across SLAM, telepresence and teleoperation, robotic manipulation, and 3D content generation. Despite their differences, these domains share common goals: photorealistic rendering, meaningful 3D structure, and accurate downstream tasks. We organize the review around unified research questions that explain why 3DGS is increasingly displacing NeRF-based approaches: What technical advantages drive its adoption? How does it adapt to different input modalities and domain-specific constraints? What limitations remain?
By systematically comparing domain-specific pipelines, we show that 3DGS balances photorealism, geometric fidelity, and computational efficiency. The survey offers a roadmap for leveraging neural rendering not only for image synthesis but also for perception, interaction, and content creation across real and virtual environments.
\end{abstract}

}

\IEEEaftertitletext{\vspace{-1.2\baselineskip}} 
\maketitle
\begin{strip}
  \centering
  \includegraphics[width=\textwidth]{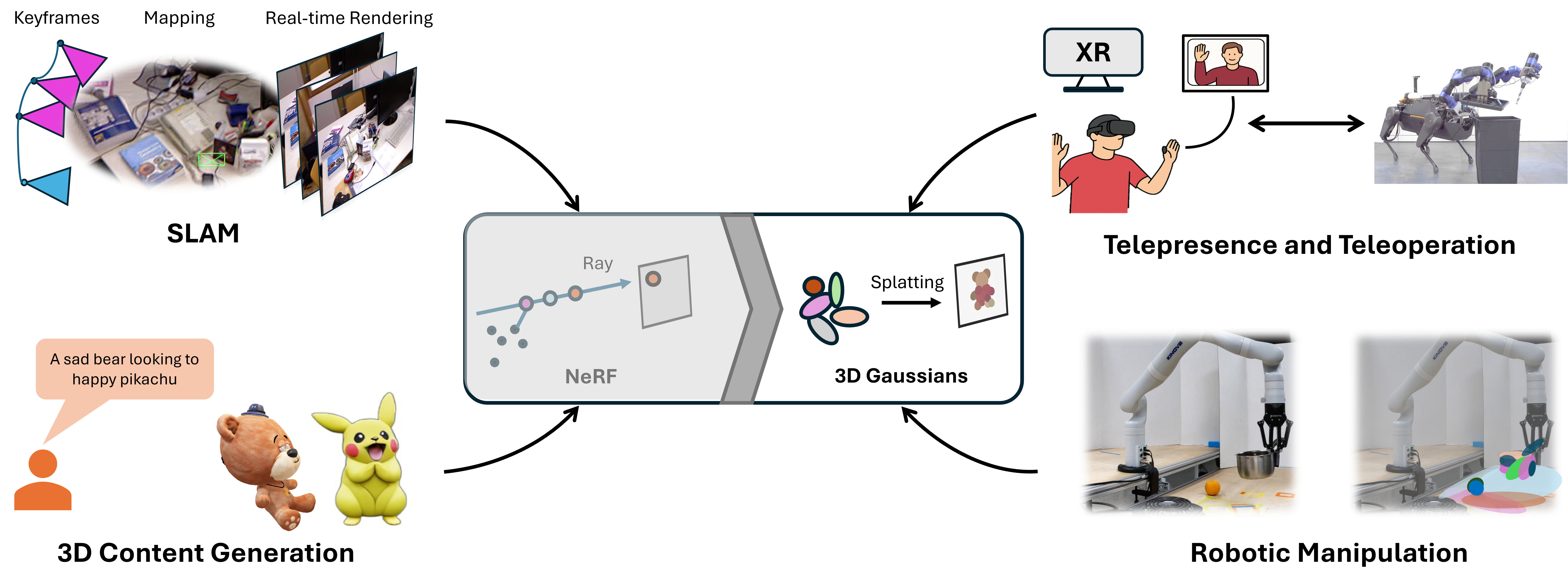}
  \captionof{figure}{\textbf{Overview.} Representative systems across four domains—SLAM, telepresence and teleoperation, 3D content generation, and robotic manipulation—use 3D Gaussian Splatting (3DGS) as a core representation. This survey examines how 3DGS is increasingly replacing traditional geometric models and NeRF-based methods, enabling unified, photorealistic, and task-adaptive scene representations.}
  \label{fig:overview}
\end{strip}

\IEEEdisplaynontitleabstractindextext

\section{Introduction}
\label{sec:intro}
\IEEEPARstart{P}{hotorealism} together with computational efficiency has become a fundamental requirement across modern 3D applications, including localization and mapping, telepresence, robotics, and 3D content generation. These applications demand scene representations that support both high-quality visualization and fast processing. Traditional 3D scene representations based on explicit geometry, such as point clouds, voxel grids, and polygonal meshes, have long been used to represent 3D environments. While effective in low-complexity settings, these approaches often struggle with novel view synthesis and detailed scene reconstruction. More recently, neural implicit representations, most notably Neural Radiance Fields (NeRF)~\cite{mildenhall2021nerf}, marked a turning point by offering view-consistent photorealism through implicit volumetric rendering. NeRF uses a multi-layer neural network to encode a scene, but its reliance on ray marching, slow convergence, and tightly coupled MLP structures limits its use in interactive systems, particularly when high-speed visualization or scene editing is required.

In this evolving landscape, 3D Gaussian Splatting (3DGS)~\cite{kerbl20233d} has emerged as a compelling middle ground between explicit and implicit approaches. By modeling scenes as spatially anchored anisotropic Gaussians and using differentiable rasterization, 3DGS enables real-time, photorealistic rendering. Unlike NeRF, which relies on optimizing neural network weights, 3DGS directly optimizes explicit Gaussian parameters while retaining differentiability, which makes it suitable for integration into optimization and learning-based pipelines.
Importantly, it offers modularity and efficient optimization, allowing integration into hybrid systems that combine classical geometry, dense prediction models, or even NeRF components. However, these advantages often come with higher memory use, because large-scale 3DGS reconstructions may require millions of primitives. 

As summarized in Fig.~\ref{fig:overview}, the rise of 3DGS reflects a broader trend toward hybrid representations that bridge the interpretability and efficiency of classical geometry with the expressiveness of neural scene models, showing adoption across domains such as dense SLAM~\cite{sandstrom2024splat}, immersive telepresence~\cite{tu2024tele}, real-time content generation~\cite{ren2023dreamgaussian4d}, and task-aware robotic perception~\cite{zheng2024gaussiangrasper3dlanguagegaussian}. 
While challenges remain, including handling extreme lighting variation, dynamic scenes, and long-term scalability, the growing body of work demonstrates that 3DGS is not merely an alternative to NeRF, but in many scenarios a versatile foundation for next-generation 3D vision and graphics systems. 

To systematically examine the rapid growth of research on neural scene representations, we conducted a structured literature review with a particular focus on 3D Gaussian Splatting. While NeRF laid the foundations for view-consistent photorealism, 3DGS has rapidly gained traction as a more efficient alternative. The primary goal of this review is to provide a comprehensive understanding of the current state of 3DGS research, its relationship to NeRF, its emerging applications, and its future potential. To ensure methodological rigor, we defined a set of guiding research questions (RQs) that structure the scope and analysis of our study: 

\begin{itemize}
    \item \textbf{RQ1}: How is 3DGS being integrated across domains like SLAM, telepresence, 3D content generation, and robotic manipulation, and what synergies emerge from these intersections? 
    \item \textbf{RQ2}: What factors motivate the transition from NeRF-based methods to 3DGS, and how do these differences manifest across application domains? 
    \item \textbf{RQ3}: What are the current technical and practical limitations preventing the widespread use of 3DGS in low-latency or real-time systems, and what alternatives or optimizations have been proposed to address these challenges? 
    \item \textbf{RQ4}: What are the most promising future research directions, and which critical application areas stand to benefit most from continued advances in 3DGS? 
\end{itemize}

\begin{figure*}[t!]
    \centering
    \includegraphics[width=\textwidth]{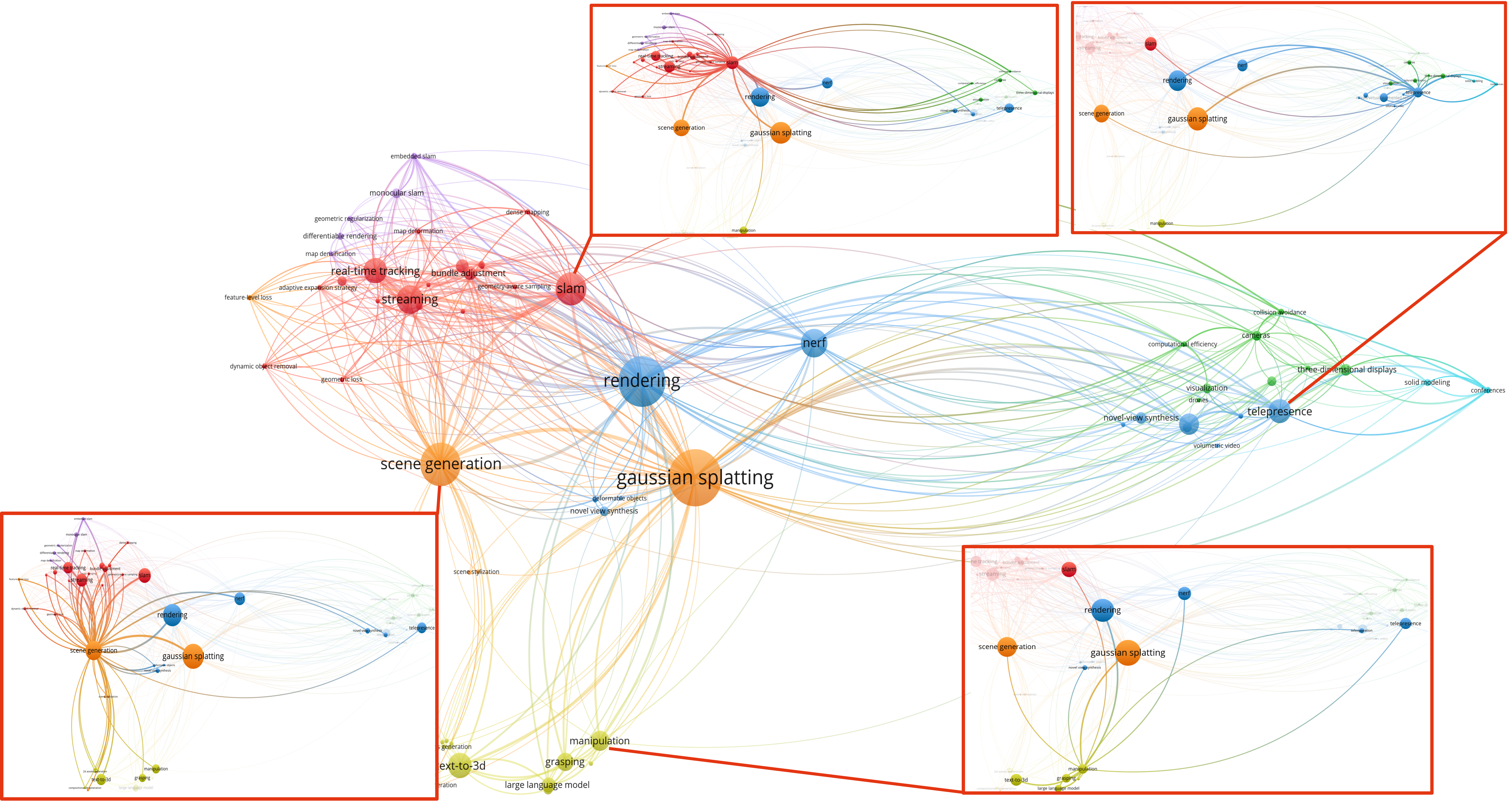}
   \caption{Keyword co-occurrence analysis using VOSviewer. Neural rendering methods such as 3DGS and NeRF are increasingly connected to downstream applications in SLAM, telepresence, 3D content generation, and robotic manipulation.}
    \label{fig:keyword_map}
\end{figure*}

Addressing these questions is crucial for consolidating fragmented progress, clarifying the comparative advantages of 3DGS over related paradigms, and informing both researchers and practitioners about its role in shaping the next generation of visual computing systems. To this end, we carried out a systematic selection of relevant literature sources. Our search encompassed major academic databases such as the ACM Digital Library, IEEE Xplore, Scopus, and Web of Science, ensuring broad coverage of peer-reviewed work. Given the fast-moving nature of this field, the search was further supplemented with recent preprints from arXiv to capture cutting-edge contributions not yet formally published. We collected a total of 140 papers published through mid-2025, spanning computer vision, robotics, and graphics venues.

To gain additional insight into the structure of the research landscape, we conducted a keyword co-occurrence analysis using VOSviewer on a curated collection of recent publications (see Fig.~\ref{fig:keyword_map}).
The results reveal an increasing frequency of co-occurrence among terms such as “Gaussian Splatting,” “NeRF,” “SLAM,” “view synthesis,” and “telepresence.” This pattern indicates that 3DGS is not developing in isolation, but rather emerging as a convergent technology intersecting with multiple established research communities in computer vision, computer graphics, and robotics. 
Motivated by these findings, this review focuses on how 3D Gaussian Splatting is being applied across four key domains: 
\begin{itemize}
    \item \textbf{SLAM:} a dense, differentiable mapping technique for real-time and high-fidelity dense mapping and tracking. 
    \item \textbf{Telepresence and Teleoperation:} immersive avatars and volumetric environments with compact streaming pipelines.
    \item \textbf{3D Content Generation:} generating high-fidelity 3D/4D objects and scenes from text or images.
    \item \textbf{Robotic Manipulation:} object-level understanding, grasp planning, and sim-to-real bridging using interpretable scene geometry.
\end{itemize}
This survey is structured as follows. Section~\ref{sec:related_and_scope} reviews prior surveys and positions our work within the broader literature, clarifying its scope and unique contributions. Section~\ref{sec:foundations} outlines the foundational models for scene representation, with a particular focus on Neural Radiance Fields (NeRF) and 3D Gaussian Splatting (3DGS), which serve as the core techniques underlying many of the applications discussed in subsequent sections. Sections~\ref{sec:slam} through~\ref{sec:manipulation} examine the adoption of these models across key domains, including Simultaneous Localization and Mapping (SLAM), immersive telepresence, 3D content generation, and robotic manipulation. Finally, Section~\ref{sec:challenges} synthesizes the open challenges identified throughout the survey and highlights promising directions for future research in this rapidly evolving field.

\section{Related Work and Survey Scope}
\label{sec:related_and_scope}
\subsection{Related Surveys}
With the rise of neural scene representations such as NeRF and 3DGS, it has become increasingly important to understand their theoretical foundation, evaluate their performance, and explore how they can be applied. Several recent surveys have examined these aspects.
For instance, \textit{3D Gaussian Splatting: A New Era}~\cite{fei2024newera3dgs} reviews the foundations of Gaussian splatting, discussing primitive modeling, differentiable rendering, and comparisons with NeRF. Complementing this, \textit{Recent Advances in 3D Gaussian Splatting}~\cite{bao20253d} analyzes optimization strategies and technical bottlenecks, highlighting issues such as real-time rendering, scalability, and visibility estimation. While these works provide solid technical grounding in rendering theory and benchmarking, their focus remains largely within computer vision, with limited attention to applications in interactive systems, simulation, or robotics.

In contrast, NeRF-focused surveys emphasize architectural evolution, deployment, and application contexts. \textit{NeRF: Neural Radiance Field in 3D Vision—A Comprehensive Review}~\cite{wang2024nerf} discusses scene understanding, perception, and motion planning in robotic systems. Similarly, \textit{Neural Radiance Fields for the Real World}~\cite{xiao2025neural} extends this to domains such as medical imaging, AR/VR, and autonomous systems. However, these surveys mainly analyze NeRF and do not cover 3DGS or its integration into broader system pipelines.

A growing body of survey literature also explores neural scene representations in specific application domains. In Simultaneous Localization and Mapping (SLAM), surveys are beginning to include both NeRF and 3DGS. For example, \textit{How NeRFs and 3DGS are Reshaping SLAM}~\cite{tosi2024nerfs} categorizes SLAM pipelines based on radiance fields and splats, studying their contributions to mapping, tracking, and relocalization. Another notable work, \textit{NeRF and 3D Gaussian Splatting SLAM in the Wild}~\cite{schmidt2024nerfgaussiansplattingslam}, evaluates outdoor SLAM scenarios under diverse conditions and highlights the advantages of 3DGS in accuracy and computational efficiency. While these works focus on SLAM-specific metrics, they do not extend their analysis to broader domains such as extended reality (XR), content creation, or robotic manipulation.

For robotics applications more broadly, Irshad et al.~\cite{irshad2024neural} provide a taxonomy of neural fields, including NeRF, 3DGS, occupancy networks, and signed distance functions (SDFs), across tasks such as manipulation, navigation, and autonomous driving. In XR and VR/AR contexts, surveys have highlighted the potential of 3DGS for high-fidelity experiences. For instance, Qiu et al.~\cite{qiu_extendReal2025} show how 3DGS enables photorealistic avatars and efficient volumetric streaming, offering a focused review of rendering pipelines and quality improvements.

Rather than detailing individual methods, we center 3DGS as the emerging core representation and organize NeRF- and geometry-based works into trend-oriented categories. This framing clarifies how research is moving toward 3DGS across SLAM, telepresence, manipulation, and 3D content generation, and it emphasizes cross-domain adoption over method-by-method catalogs. 

\subsection{Scope of This Survey}
This survey provides a comprehensive overview of neural scene representations, with a particular emphasis on NeRF and 3DGS. We begin by outlining the foundational components that underpin modern neural scene representations, such as structure-from-motion (SfM) and feed-forward methods, which supply essential parameters for NeRF and 3DGS. 
We then position NeRF and 3DGS as the central building blocks that enable a wide range of downstream applications.

Unlike prior surveys that emphasize the architectural details of individual methods, our focus is on the broader research landscape: how representative methods align with emerging trends, what tasks they address, and how they compare against alternative approaches. To support this perspective, we provide comparative tables with numerical evaluations and unified figures that categorize methods across domains, complemented by representative visual examples. This approach highlights application-level insights and cross-domain positioning, rather than repeating architectural descriptions that are already covered in original research papers.

The key contributions of this survey are as follows:
\begin{itemize}
    \item We present a systematic comparison of NeRF and 3DGS, emphasizing their role within application trends and their suitability across domains such as robotics, SLAM, XR, and interactive content creation. 
    \item We categorize representative methods within each application domain, analyze the tasks they perform, and compare their contributions through both tabular and visual summaries.
    \item Based on these structured comparisons, we outline open research directions.
\end{itemize}

\section{Foundational Components}
\label{sec:foundations}

Modern neural scene representations such as Neural Radiance Fields (NeRF) and 3D Gaussian Splatting (3DGS) are deeply rooted in decades of advances in computer vision, geometry, and rendering.
To understand their impact on real-time 3D systems, it is important to briefly review the building blocks that support them.

In this section, we first review the progression of scene reconstruction techniques, spanning from classical geometry-based pipelines to feedforward learning-based predictors, and analyze how these approaches provide the essential backbone for NeRF and 3DGS. 
We then contrast implicit radiance-field formulations with explicit 3D Gaussian-splatting representations, highlighting the trade-offs between flexibility, efficiency, and fidelity. This comparison illustrates why 3DGS has emerged as a compelling foundation for real-time applications.

\subsection{Scene Reconstruction Methods}

\subsubsection{Traditional Methods}
3D geometry forms the backbone of SLAM, content generation, and manipulation pipelines, providing depth, geometry, and camera poses for downstream tasks. Traditionally, Structure-from-Motion (SfM)~\cite{oliensis2000critique, nocerino20173d, tefera20183dnow} was designed to estimate camera parameters and depth maps from multiview images. Research such as~\cite{liu2024robust, Schonberger_2016_CVPR, snavely2006photo, wu2013towards} advanced SfM through feature detection, image matching, triangulation, and bundle adjustment, with widely used systems including COLMAP~\cite{Schonberger_2016_CVPR} and GLOMAP~\cite{pan2024glomap}. 

Recent advances have replaced hand-engineered modules with deep learning, improving robustness and scalability. Learned keypoint detectors~\cite{detone2018superpoint, dusmanu2019d2, tyszkiewicz2020disk, yi2016lift} and feature matchers~\cite{chen2021learning, sarlin2020superglue, shi2022clustergnn} enable hybrid or differentiable SfM systems such as VGGSfM~\cite{wang2024vggsfm}, BA-Net~\cite{tang2018ba}, DeepV2D~\cite{teed2018deepv2d}, Deep Two-View SfM~\cite{wang2021deep}, and DeepSfM~\cite{wei2020deepsfm}. These methods outperform classical SfM in challenging environments but still face scalability limitations in interactive settings.

\subsubsection{Feedforward Scene Reconstruction}
Feedforward methods directly predict dense scene geometry from images, often bypassing explicit calibration. DUSt3R~\cite{Wang_2024_CVPR} and MASt3R~\cite{leory_mast3r} predict dense point clouds and metric-scale poses from image pairs. MASt3R-SfM~\cite{duisterhof2024mast3r} extends this to multiview global optimization, though long sequences remain difficult to scale. Other approaches such as CUT3R~\cite{wang2025cut3r}, Fast3R~\cite{yang2025fast3r}, FLARE~\cite{zhang2025FLARE}, and Spann3R~\cite{wang20253d_spann3r} improve reconstruction efficiency, while temporal models like MonST3R~\cite{zhang2024monst3r} support online SLAM. Diffusion-based models such as Geo4D~\cite{jiang2025geo4d} and Aether~\cite{team2025aether} extend to uncertain or sparse inputs, including 4D dynamics. Pow3R~\cite{jang2025pow3r} generalizes DUSt3R by optionally accepting intrinsics, poses, and depth, improving flexibility and reconstruction accuracy.

Some prediction techniques can output Gaussian parameters directly. Splatt3R~\cite{smart2024splatt3r} estimates splatting primitives from two views, while PreF3R~\cite{chen2024pref3r} extends this to multiview inputs. Reloc3R~\cite{dong2025reloc3r} adapts DUSt3R for pose regression and large-scale localization, and VGGT~\cite{wang2025vggt} performs multiframe reconstruction with direct pose estimation, making it attractive for initializing 3D Gaussians efficiently.

These techniques map image collections to geometry, pose, and confidence as follows:
\[
f_\theta: \{I_1, I_2, \dots, I_N\} \rightarrow \{X_i, D_i, P_i, \Sigma_i\}_{i=1}^{N},
\]
where \(X_i\) are point clouds, \(D_i\) depth maps, \(P_i\) camera poses, and \(\Sigma_i\) uncertainty estimates. Most use vision transformers with fusion and regression heads~\cite{Ranftl_2021_ICCV}:
\[
H = \mathrm{Fusion}(\{\phi(I_i)\}), \quad D_i = \psi_{\text{depth}}(H_i), \quad P_i = \psi_{\text{pose}}(H_i).
\]

Training objectives combine geometric regression, multiview consistency, and uncertainty weighting. Dense predictors increasingly serve as the geometric front-end for NeRF and 3DGS. For NeRF, they supply initial poses, depth priors, or sparse geometry to bootstrap radiance-field optimization. For 3DGS, they are even more critical, providing initialization, refinement, and supervision for splat parameters. Their real-time capabilities make them indispensable in dynamic pipelines. In summary, dense reconstruction bridges classical geometry and modern neural rendering, providing the backbone for real-time, photorealistic applications.

\subsection{Neural Radiance Fields (NeRF)}

Neural Radiance Fields (NeRF)~\cite{mildenhall2021nerf} introduced a major paradigm shift in scene representation and novel view synthesis. By learning a continuous volumetric function from sparse 2D inputs, NeRF achieves dense, view-consistent, and photorealistic rendering. Despite its impact, NeRF’s reliance on volumetric ray marching and slow convergence makes it less suitable for real-time systems.

NeRF is defined as a neural function:
\[
\mathcal{F}_\theta: (\mathbf{x}, \mathbf{d}) \mapsto (\mathbf{c}, \sigma),
\]
where \( \mathbf{x} \in \mathbb{R}^3 \) is a 3D point, \( \mathbf{d} \in \mathbb{S}^2 \) is a viewing direction, \( \mathbf{c} \in \mathbb{R}^3 \) is the emitted RGB color, and \( \sigma \in \mathbb{R}^+ \) is the volumetric density.

To render a scene, a camera ray \( \mathbf{r}(t) = \mathbf{o} + t\mathbf{d} \) is projected, and color is obtained by alpha compositing along the ray:
\[
C(\mathbf{r}) = \int_{t_n}^{t_f} T(t)\, \sigma(\mathbf{r}(t))\, \mathbf{c}(\mathbf{r}(t), \mathbf{d})\, dt,
\]
with transmittance
\[
T(t) = \exp\!\left(-\int_{t_n}^{t}\sigma(\mathbf{r}(s))\, ds\right).
\] 

In practice, this is approximated by stratified sampling with \(N\) intervals of size \(\delta_i\):
\[
C(\mathbf{r}) \approx \sum_{i=1}^{N} T_i \Big(1 - e^{-\sigma_i \delta_i}\Big) \mathbf{c}_i,
\]
where \(T_i = \prod_{j<i} e^{-\sigma_j \delta_j}\) is the accumulated transmittance.

During training, the model is optimized using a photometric loss between predicted and ground-truth pixels:
\[
\mathcal{L}_{\text{NeRF}} = \sum_{\mathbf{r} \in \mathcal{R}} \big\| C(\mathbf{r}) - \hat{C}(\mathbf{r}) \big\|_2^2.
\]

NeRF produces compact, differentiable, and photorealistic representations that are well suited for offline rendering. However, it faces several limitations:  
\begin{itemize}
    \item It requires dense sampling and expensive ray marching, which limit speed.  
    \item Convergence is slow, often requiring minutes to hours of optimization.  
    \item It depends strongly on accurate camera poses.  
    \item Scalability to large or dynamic scenes remains limited.  
\end{itemize}

\subsection{3D Gaussian Splatting (3DGS)}
3D Gaussian Splatting (3DGS)~\cite{kerbl20233d} presents a further shift in real-time neural rendering. Unlike NeRF’s implicit volumetric fields, 3DGS models scenes explicitly as a set of anisotropic Gaussians. Differentiable rasterization enables interactive frame rates without sacrificing photorealism, which makes 3DGS particularly suited for dynamic and interactive systems.

A Gaussian primitive \( \mathcal{G}_i \) is defined by:
\begin{itemize}
    \item Center: \( \mu_i \in \mathbb{R}^3 \),  
    \item Covariance: \( \Sigma_i \in \mathbb{R}^{3 \times 3} \), positive semidefinite,  
    \item Opacity: \( \alpha_i \in [0,1] \),  
    \item Color: \( c_i \in \mathbb{R}^3 \), often with view-dependent shading (e.g., spherical harmonics).  
\end{itemize}

A Gaussian scene is projected and rendered into image space by linearizing camera projection \( \pi(W\mu_i) \) with Jacobian \(J\):  
\[
\Sigma'_i = J W \Sigma_i W^\top J^\top, 
\quad \mu'_i = \pi(W\mu_i),
\]
where \(W\) is the world-to-camera transform.  
Pixels are rendered using ordered alpha compositing:
\[
C(x) = \sum_{n=1}^{N} c_n \alpha'_n \prod_{j=1}^{n-1}(1-\alpha'_j),
\]
with effective opacity:
\[
\alpha'_n = \alpha_n \exp\!\Big(-\tfrac{1}{2}(x-\mu'_n)^\top \Sigma_n'^{-1}(x-\mu'_n)\Big).
\]

Primitive parameters \( \{\mu_i, \Sigma_i, \alpha_i, c_i\} \) are optimized via image-reconstruction losses:
\[
\mathcal{L} = (1-\lambda)\,\mathcal{L}_{\text{L1}} + \lambda\,\mathcal{L}_{\text{DSSIM}}.
\]
To ensure valid covariances, \(\Sigma_i\) is reparameterized as:
\[
\Sigma_i = R_i S_i S_i^\top R_i^\top, \quad R_i \in SO(3),\ S_i=\text{diag}(s_i).
\]

Densification and pruning adapt the number of Gaussians to scene structure. Primitives can be cloned, split, or pruned based on gradient magnitude and coverage, which enables scalability and adaptivity in real-time rendering.

Gaussian-based representations offer several benefits for practical 3D applications:
\begin{itemize}
    \item Real-time rendering can exceed 130~FPS on commodity GPUs.  
    \item Compared with dense volumetric grids, Gaussian primitives provide an explicit encoding of 3D geometry and appearance, offering high visual fidelity with efficient rendering.  
\end{itemize}

In summary, while NeRF introduced continuous volumetric rendering, 3DGS establishes an explicit, efficient, and real-time alternative that is increasingly adopted as the backbone for interactive 3D vision and graphics systems.

\begin{figure*}[ht]
\centering
\includegraphics[width=\linewidth]{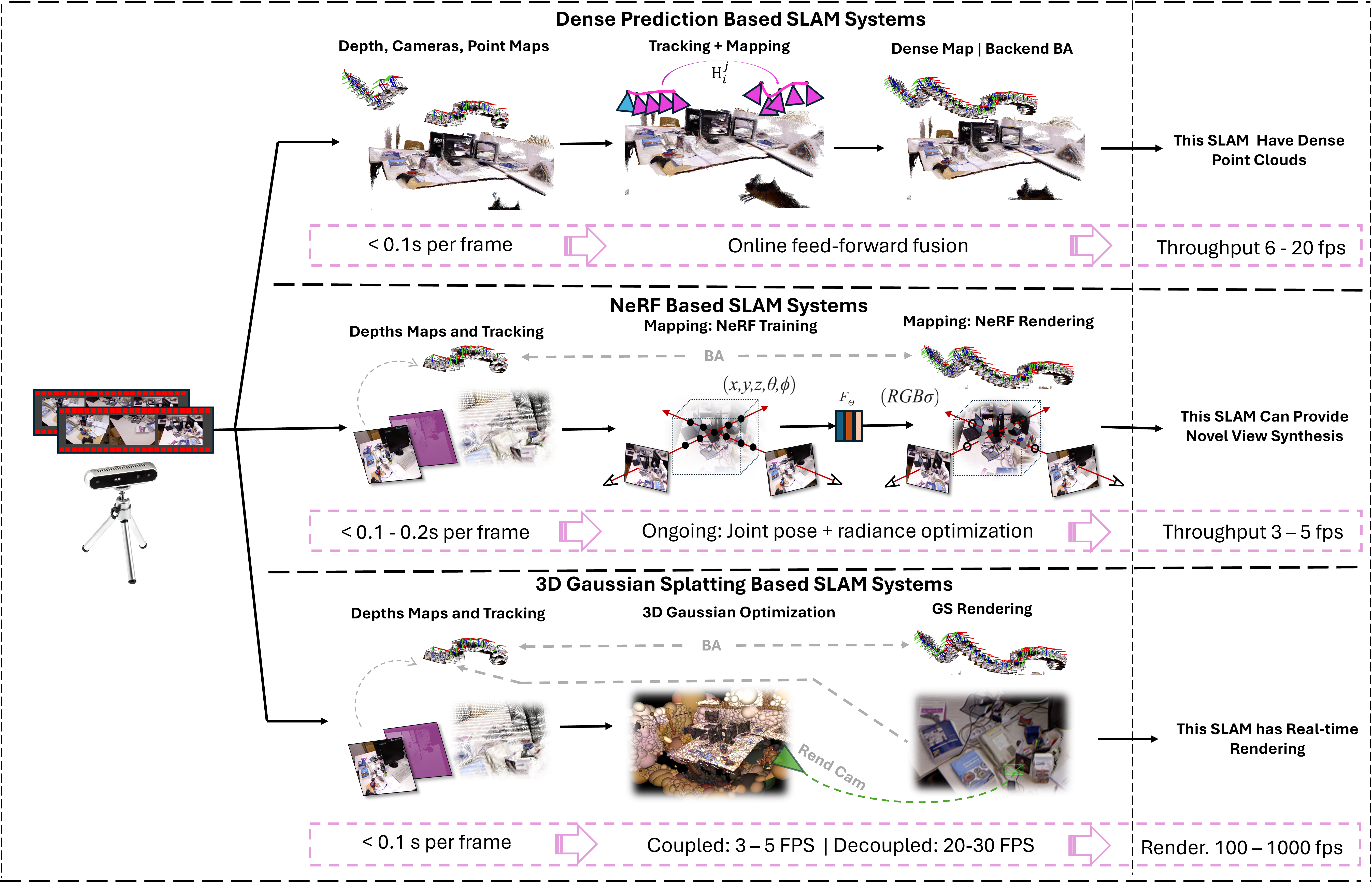}
\caption{Representative pipelines across three families of SLAM systems.
Top: dense-prediction SLAM (e.g., MASt3R-SLAM~\cite{murai2024mast3r_slam}, VGGT-SLAM~\cite{maggio2025vggt}) leverages feedforward geometric priors for rapid depth and pose inference, typically achieving 6--20~FPS throughput.
Middle: NeRF-based SLAM (e.g., NICE-SLAM~\cite{Zhu_2022_CVPR}, GO-SLAM~\cite{Zhang_2023_ICCV}) jointly optimizes poses and radiance fields, enabling dense, photorealistic maps but limited to 3--15~FPS due to volumetric ray marching.
Bottom: 3DGS-based SLAM (e.g., MonoGS~\cite{Matsuki_2024_CVPR}, GS-ICP-SLAM~\cite{ha2024rgbd}) uses explicit Gaussian primitives with rasterization. Coupled designs reach 3--5~FPS, while decoupled pipelines sustain 20--30~FPS SLAM throughput and support rendering speeds exceeding 100~FPS and reaching up to 1000~FPS. Overall, the figure illustrates the evolution from feedforward predictors, through implicit NeRF optimizers, to explicit 3DGS as a practical foundation for real-time, photorealistic SLAM.}
\label{fig:SLAM_main_fig}
\end{figure*}

\begin{table*}[ht]
\centering
\begin{tabular}{|l|l|l|l|c|c|}
\hline
\textbf{Paper} & \textbf{Type} & \textbf{Input} & \textbf{Task} & \textbf{Year} & \textbf{FPS} \\
\hline
\multicolumn{6}{|c|}{\textbf{Dense Prediction-based SLAM}} \\
\hline
MASt3R-SLAM~\cite{murai2024mast3r_slam} & Dense pred. & Mono & Feedforward point pred. + dense SLAM & 2025 & \textdagger~15 \\
VGGT-SLAM~\cite{maggio2025vggt} & Dense pred. & Mono (MV) & \(\mathrm{SL}(4)\) align.; xformer-based dense mapping & 2025 & --- \\
\hline
\multicolumn{6}{|c|}{\textbf{NeRF-based SLAM}} \\
\hline
NICE-SLAM~\cite{Zhu_2022_CVPR} & NeRF-based & RGB-D & Online NeRF optim.; hash features & 2022 & --- \\
GO-SLAM~\cite{Zhang_2023_ICCV} & NeRF-based & Mono / Stereo / RGB-D & Loop closure + global BA & 2023 & \textdagger~8 \\
GLORIE-SLAM~\cite{zhang2024glorie} & NeRF-based (neural points) & Mono & DSPO + global optim. & 2024 & --- \\
\hline
\multicolumn{6}{|c|}{\textbf{3D Gaussian Splatting SLAM (Coupled)}} \\
\hline
MonoGS~\cite{Matsuki_2024_CVPR} & GS (coupled) & Mono & Joint pose + Gaussian optim. & 2024 & \textdagger~3 \\
SplaTAM~\cite{Keetha_2024_CVPR} & GS (coupled) & RGB-D & Pose + map co-optim. via GS rasterization & 2024 & 400 \\
LoopSplat~\cite{zhu2024loopsplat} & GS (coupled) & Mono / RGB-D & Splat-based LC + pose correction & 2025 & --- \\
\hline
\multicolumn{6}{|c|}{\textbf{3D Gaussian Splatting SLAM (Decoupled)}} \\
\hline
Photo-SLAM~\cite{huang2024photo} & GS (decoupled) & Mono / Stereo / RGB-D & ORB-SLAM3 tracking + GS mapping & 2024 & 1000 \\
GS-ICP-SLAM~\cite{ha2024rgbd} & GS (decoupled) & RGB-D & G-ICP tracking + GS rendering & 2024 & 107 \\
RTG-SLAM~\cite{pengRTG-SLAM} & GS (decoupled) & RGB-D & Frame-to-model ICP + efficient GS optim. & 2024 & \textdagger~17.9 \\
IG-SLAM~\cite{sarikamis2024ig} & GS (decoupled) & RGB & BA tracking + proxy-depth–guided GS & 2024 & \textdagger~10 \\
CaRtGS~\cite{feng2025cartgs} & GS (decoupled) & RGB-D & Adaptive KFs + splat-wise backprop. & 2025 & --- \\
GS-SLAM~\cite{Yan_2024_CVPR} & GS (decoupled) & RGB-D & Coarse-to-fine + BA + Gaussian fusion & 2024 & 386 \\
SemGauss-SLAM~\cite{zhu2024semgauss} & GS (decoupled) & RGB-D & Joint pose/geom./sem. optim. & 2025 & --- \\
Compact3DGS-SLAM~\cite{deng2024compact} & GS (decoupled) & RGB-D & Wnd. masking + compact geom. cbk & 2024 & --- \\
\hline
\multicolumn{6}{|c|}{\textbf{Hybrid / Advanced / Navigation}} \\
\hline
DenseSplat~\cite{li2025densesplat} & Hybrid (GS+NeRF) & RGB-D & Sparse KFs + NeRF densif. + GS BA & 2025 & --- \\
GARAD-SLAM~\cite{li2025garad} & Dynamic GS-SLAM & RGB-D & Gaussian-level dynamic segm. + filtering & 2025 & --- \\
MAGiC-SLAM~\cite{yugay2024magic} & Multiagent GS-SLAM & RGB-D (multi-agent) & Joint mapping + central optim. & 2024 & --- \\
Splat-Nav~\cite{chen_splat-nav} & Robotic GS maps & RGB & RGB-only tracking + replanning & 2024 & \textdagger~25 \\
\hline
\end{tabular}
\caption{Comparison of representative SLAM systems. FPS refers to \emph{rendering throughput} when explicitly reported; entries marked \textdagger~indicate \emph{end-to-end} SLAM runtime. Abbrev.: BA—bundle adjustment; DSPO—differentiable sparse direct pose optim.; FPS—frames per second; geom.—geometry; G-ICP—generalized ICP; GS—Gaussian splatting; KFs—keyframes; LC—loop closure; Mono—monocular; MV—multiview; NeRF—Neural Radiance Fields; optim.—optimization; RGB-D—RGB plus depth; xformer—transformer; cbk—codebook.}
\label{tab:SLAM_main_table}
\end{table*}

\section{SLAM: Toward High-Fidelity 3D Gaussian Splatting}
\label{sec:slam}

In the previous section, we examined emerging research paradigms that adapt Gaussian representations and NeRF for 3D reconstruction. This raises the question: Why are modern SLAM systems shifting toward 3D Gaussian Splatting (3DGS)? Classical SLAM provides reliable geometry and real-time tracking, but its maps are typically sparse and lack visual fidelity. NeRF-based SLAM improves photorealism but suffers from heavy volumetric rendering and low frame rates. In contrast, 3DGS offers an explicit, differentiable, and rasterization-friendly representation, bridging the gap between photorealistic mapping and interactive performance. In this section, we trace this evolution and clarify the roles of dense prediction, NeRF, and 3DGS within modern SLAM pipelines.

As shown in Fig.~\ref{fig:SLAM_main_fig}, representative pipelines highlight the evolution from dense prediction to NeRF- and 3DGS-based SLAM, illustrating the trade-offs at the module level. Table~\ref{tab:SLAM_main_table} summarizes FPS across paradigms: dense-prediction SLAM systems sustain 6--20~FPS throughput, NeRF-based methods achieve 3--15~FPS, and 3DGS-based methods range from 3--30~FPS in end-to-end SLAM and exceed 100~FPS (up to 1000~FPS) in rendering.

\subsection{Feedforward Network--Based SLAM}
Dense SLAM systems increasingly rely on feedforward predictors for depth, points, or poses as front ends, often decoupling tracking from mapping.
MASt3R-SLAM~\cite{murai2024mast3r_slam} integrates a metric-scale two-view prior and estimates poses via RANSAC on predicted pointmaps, refining with photometric consistency (\(\sim\)15~FPS).
VGGT-SLAM~\cite{maggio2025vggt} fuses multiframe transformer features and resolves projective ambiguity by optimizing over \(\mathrm{SL}(4)\), producing globally aligned dense reconstructions, though at significant memory cost.

Most of these techniques follow a common formulation. Let \(\mathcal{I}=\{I_t\}_{t=1}^T\). Dense predictors render depth or pointmaps \(D_t, X_t = f_\theta(I_t)\). Poses are estimated either from robust alignment
\[
T_{t-1,t}=\arg\min_{T\in \mathrm{SE}(3)}\sum_{u}\big\|T\cdot P_t(u)-P_{t-1}(\hat u)\big\|^2
\]
or from a learned head \(T_t=\psi_{\text{pose}}(H_t)\). Local maps are fused into a global model via pose-graph optimization or \(\mathrm{SL}(4)\) alignment,
\(\hat P_i=H_i P_i,\ H_i\in \mathrm{SL}(4)\), and finally merged into a point or splat map \(\mathcal{M}=\mathrm{Fuse}(\{H_t P_t\})\).

\subsection{NeRF-Based SLAM}
NeRFs~\cite{mildenhall2021nerf} provide continuous, view-consistent mapping but require dense ray marching.
Early works such as iNeRF~\cite{yen2021inerf} demonstrated single-frame pose recovery, while
iMAP~\cite{sucar2021imap} established joint pose and radiance optimization from RGB-D.
NICE-SLAM~\cite{Zhu_2022_CVPR} introduced hash-encoded grids and keyframe scheduling.
GO-SLAM~\cite{Zhang_2023_ICCV} added online loop closure and full bundle adjustment (BA) but still runs at about 8~FPS.
GLORIE-SLAM~\cite{zhang2024glorie} replaces volumetric fields with neural points and differentiable sparse direct pose optimization, improving scalability without per-ray MLPs.
In practice, NeRF-based SLAM offers strong fidelity but struggles with latency and incremental updates.

\subsection{3DGS-Based SLAM}
MonoGS~\cite{Matsuki_2024_CVPR} is a monocular 3DGS SLAM that jointly optimizes poses and Gaussians at around 3~FPS (end-to-end).
SplaTAM~\cite{Keetha_2024_CVPR} co-optimizes tracking and mapping with a GS rasterizer and demonstrates about 400~FPS rendering.
LoopSplat~\cite{zhu2024loopsplat} adds splat-based loop closure for global consistency.
Photo-SLAM~\cite{huang2024photo} combines ORB-SLAM3 tracking with GS mapping and reports up to 1000~FPS rendering.
GS-ICP-SLAM~\cite{ha2024rgbd} aligns G-ICP tracking with a GS map, reaching 107~FPS rendering.
RTG-SLAM~\cite{pengRTG-SLAM} stabilizes Gaussian updates with partial-pixel optimization, achieving 17.9~FPS end-to-end.
IG-SLAM~\cite{sarikamis2024ig} uses dense BA and proxy depth for updates (about 10~FPS end-to-end).
GS-SLAM~\cite{Yan_2024_CVPR} employs adaptive expansion and coarse-to-fine tracking, with 386~FPS rendering.

Recent works extend 3DGS SLAM beyond static, single-agent reconstruction to address hybrid, dynamic, and multiagent scenarios. DenseSplat~\cite{li2025densesplat} densifies GS with NeRF priors.
GARAD-SLAM~\cite{li2025garad} segments dynamic Gaussians.
MAGiC-SLAM~\cite{yugay2024magic} merges multiagent submaps.
Splat-Nav~\cite{chen_splat-nav} employs GS maps for planning and RGB-only localization at around 25~Hz.

Dense-prediction front ends reduce the burden of data association, while NeRF-based SLAM improves visual fidelity at the cost of high latency. In contrast, 3D Gaussian Splatting delivers the fastest rendering speeds while maintaining competitive accuracy, enabling photorealistic and interactive SLAM. Remaining challenges include achieving reliable loop closure at the splat level, improving memory efficiency and compactness, and ensuring robustness in textureless or highly dynamic environments.

\begin{figure*}[t]
    \centering
    \includegraphics[width=\linewidth]{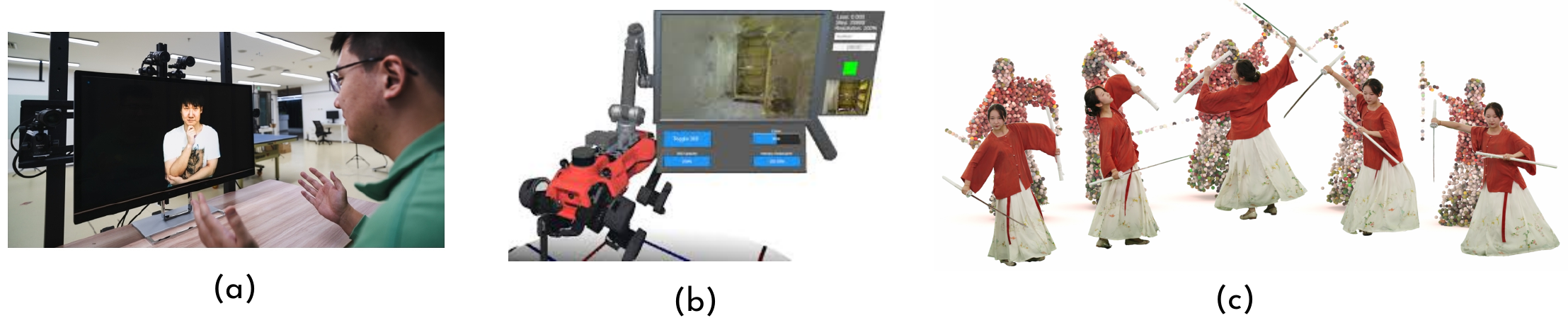}
    \caption{Illustrative applications of Gaussian splatting in telepresence and teleoperation: (a) telepresence systems, (b) robotic manipulation, and (c) 3D scene reconstruction.}
    \label{fig:telepresenceSample}
\end{figure*}

\section{Telepresence and Teleoperation}
\label{sec:telepresence}
Telepresence is a form of communication that enables individuals in two or more geographically separated spaces, such as offices, lounges, or meeting rooms, to interact as if they were in the same physical environment~\cite{oyamarobots}. As illustrated in Fig.~\ref{fig:telepresenceSample}, TeleAloha~\cite{tu2024tele} presents a telepresence system between two individuals located in separate physical spaces, allowing participants to see and hear each other in real time. The experience is designed to be immersive, with life-sized three-dimensional visuals and synchronized audio that create a strong sense of co-presence despite the distance.

Teleoperation (also referred to as telerobotics) involves the remote control of robotic systems to carry out tasks~\cite{oyamarobots}. It is especially important for complex operations in challenging, dangerous, or otherwise inaccessible environments, such as deep-sea exploration, space missions, nuclear facilities, and disaster zones. As depicted in Fig.~\ref{fig:telepresenceSample} (from~\cite{patil2024radiance}), teleoperation enables human operators to manipulate robotic devices from a safe distance, often using precision tools, cameras, and sensors that provide real-time feedback to guide the robot’s actions.

\subsection{Requirements for Realistic Telepresence}
An ideal telepresence and telerobotics interface should meet several key requirements to deliver a realistic and immersive user experience~\cite{oyamarobots,tef_tele,tefera2023towards}. These capabilities are critical:
\begin{enumerate}
    \item \textbf{Multi-user immersive visualization:} Provide correct physical scale and appropriate viewpoints to support life-sized, three-dimensional views. Multiple users should be able to view and interact with the remote environment while preserving spatial scale and depth perception for a natural experience~\cite{fuchs2014immersive}.
    \item \textbf{Accurate dynamic geometric modeling:} Maintain a highly accurate geometric model of the remote environment and any people within it~\cite{fuchs2014immersive}. The model should update dynamically to reflect motion, gestures, and object manipulation.
    \item \textbf{Real-time environmental scanning and updates:} Continuously scan the remote space to maintain an up-to-date 3D representation~\cite{tefera2023towards}. Latency should be minimized to keep user actions and system feedback aligned.
    \item \textbf{Photorealistic, visually engaging output:} Deliver realistic lighting, shading, color, and texture to enhance immersion and believability.
\end{enumerate}

While achieving all of these requirements is highly challenging, researchers have made significant progress by focusing on specific subsets of these capabilities. Instead of aiming for a fully integrated, real-time, photorealistic system, many studies have explored more attainable approaches—such as reconstructing and streaming digital avatars of users to convey presence, or developing systems dedicated solely to remote robotic operation when physical intervention is not feasible or safe.

\begin{table*}[htbp]
\centering
\begin{tabular}{|l|l|l|l|c|c|}
\hline
\textbf{Paper} & \textbf{Type} & \textbf{Input} & \textbf{Task} & \textbf{Year} & \textbf{FPS} \\
\hline
\multicolumn{6}{|c|}{\textbf{Volumetric Streaming}} \\
\hline
SwinGS~\cite{liu2024swings} & GS & --- & Volumetric video streaming & arXiv 2023 & 300+ \\
\hline
DualGS~\cite{jiang2024robust} & GS & --- & Volumetric video playback & TOG 2024 & 77+ \\
\hline
\cite{wu2025advancing} & GS & --- & Volumetric video streaming & IWMCSA 2025 & 45 \\
\hline
\multicolumn{6}{|c|}{\textbf{Immersive Content}} \\
\hline
VR-GS~\cite{jiang2024vr} & GS & --- & Generation and editing & SIGGRAPH 2024 & 135+ \\
\hline
VR-Splatting~\cite{franke2024vr} & GS & --- & VR rendering & arXiv 2025 & --- \\
\hline
ExScene~\cite{gong2025exscene} & GS & --- & Single-view 3D reconstruction & arXiv 2025 & --- \\
\hline
DASS~\cite{liu2024dynamics} & GS & --- & Dynamic, real-time 4D scene reconstruction & arXiv 2025 & 254 \\
\hline
\multicolumn{6}{|c|}{\textbf{Telepresence and Teleoperation}} \\
\hline
\cite{patil2024radiance} & NeRF+GS & --- & Teleoperation & IROS 2024 & NeRF (0.980~FPS), GS (151~FPS) \\
\hline
Reality Fusion~\cite{li2024reality} & GS & --- & Teleoperation & IROS 2024 & 30--35~FPS \\
\hline
TeleAloha~\cite{tu2024tele} & GS & --- & Telepresence & SIGGRAPH 2024 & 30 \\
\hline
\cite{cao2025real} & GS & --- & Telerehabilitation & TVCG 2025 & 400~FPS \\
\hline
\end{tabular}
\caption{Comparison of differentiable rendering methods in telepresence and teleoperation.}
\label{tab:telepresence}
\end{table*}

\subsection{Real-Time Neural Avatars and Streaming}
One notable example is TeleAloha by Tu~\emph{et~al.}~\cite{tu2024tele}, a bidirectional telepresence system for peer-to-peer communication. Using four sparse RGB cameras, a consumer-grade GPU, and an autostereoscopic screen, it achieves 2{,}048\,$\times$\,2{,}048 resolution at 30~FPS with end-to-end latency under 150~ms. The paper introduces a view-synthesis pipeline for upper-body representation that combines a cascaded disparity estimator for robust geometry cues with a neural rasterizer based on Gaussian splatting; a weighted blending mechanism refines the output to 2{,}048\,$\times$\,2{,}048 resolution.

In a related domain, Cao~\emph{et~al.}~\cite{cao2025real} propose a data-driven approach for telerehabilitation that uses a single RGB camera and 3D Gaussian splatting to generate real-time, photorealistic, free-viewpoint renderings of patients. By retargeting dynamic human motion to a canonical pose, their system achieves accurate 3D reconstruction while addressing real-world challenges such as occlusions from exercise equipment.

DualGS~\cite{jiang2024robust} presents a Gaussian-based approach for real-time, high-fidelity volumetric video playback. It disentangles motion and appearance, achieving up to \(120\times\) compression, enabling immersive VR experiences with minimal storage requirements and enhancing user engagement.

SwinGS~\cite{liu2024swings} introduces a framework that adapts 3D Gaussian splatting (3DGS) for real-time volumetric video streaming, addressing challenges like excessive model sizes and content drift. It uses a sliding-window technique to capture Gaussian snapshots across frames, enabling efficient streaming of long videos. SwinGS integrates spacetime Gaussians with Markov chain Monte Carlo (MCMC) for dynamic scene adaptation, achieving an 83.6\% reduction in transmission cost while maintaining quality.

Liu~\emph{et~al.}~\cite{liu2024dynamics} propose a dynamic, real-time 4D scene reconstruction framework that advances streaming-based volumetric rendering. Unlike approaches that require full multiview sequences before processing, it allows frame-by-frame reconstruction for live applications such as telepresence, AR/VR, and robotics. The system operates through a three-stage pipeline: selective inheritance carries forward relevant Gaussian primitives to maintain temporal coherence; dynamics-aware shift distinguishes static and moving elements; and error-guided densification adds new Gaussians in regions with high reconstruction error.

\subsection{Applications in Remote Collaboration and Robotics}
High-quality visual feedback plays a critical role in enhancing an operator’s situational awareness, which is essential for effective decision making and control in remote environments. However, achieving such fidelity often increases memory and bandwidth requirements, posing challenges for real-time deployment.

Researchers have applied Neural Radiance Fields (NeRF) and 3D Gaussian splatting (3DGS) to enhance remote collaboration and robotic teleoperation, aiming to deliver high-fidelity, interactive scene representations at real-time rates. Patil~\emph{et~al.}~\cite{patil2024radiance} leverage radiance-field methods, including NeRF and 3DGS, for maneuverable scene reconstructions with online training from live sensor data, enabling dynamic adaptation to new environments. Integrated VR visualization improves situational awareness and control, and the framework supports multiple radiance-field variants for flexibility.

RealityFusion by Li~\emph{et~al.}~\cite{li2024reality} focuses on teleoperation via a VR interface powered by RGB-D sensing and 3DGS. The system enables seamless switching between egocentric and exocentric viewpoints, improving navigation and spatial understanding. Implemented within a Unity-based VR environment, it allows intuitive interaction with standard VR headsets and controllers. User studies indicate improved task performance, usability, and satisfaction compared with point-cloud baselines.

Beyond teleoperation, several studies explore 3DGS for immersive content creation and manipulation in VR. VR-GS~\cite{jiang2024vr} introduces physically aware Gaussian splatting, real-time deformation embedding, and dynamic shadow rendering to enhance user interaction. VR-Splatting~\cite{franke2024vr} employs foveated rendering—neural point rendering for the fovea and 3DGS for the periphery—to balance fidelity and efficiency.

Expanding to 3D scene reconstruction, ExScene~\cite{gong2025exscene} proposes a two-stage pipeline from a single image: a multimodal diffusion model synthesizes a panoramic image, a panoramic depth estimator provides geometry, and a refinement stage applies video-diffusion priors with camera-trajectory consistency and color–geometry coherence to fine-tune the 3DGS output.

Together, these works highlight the potential of these methods—especially 3D Gaussian splatting—for real-time, high-fidelity interaction and visualization across VR, teleoperation, and scene-reconstruction applications.

\section{3D Content Generation}
\label{sec:AIGC}
\begin{figure*}[h!]
    \centering
    \includegraphics[width=\linewidth]{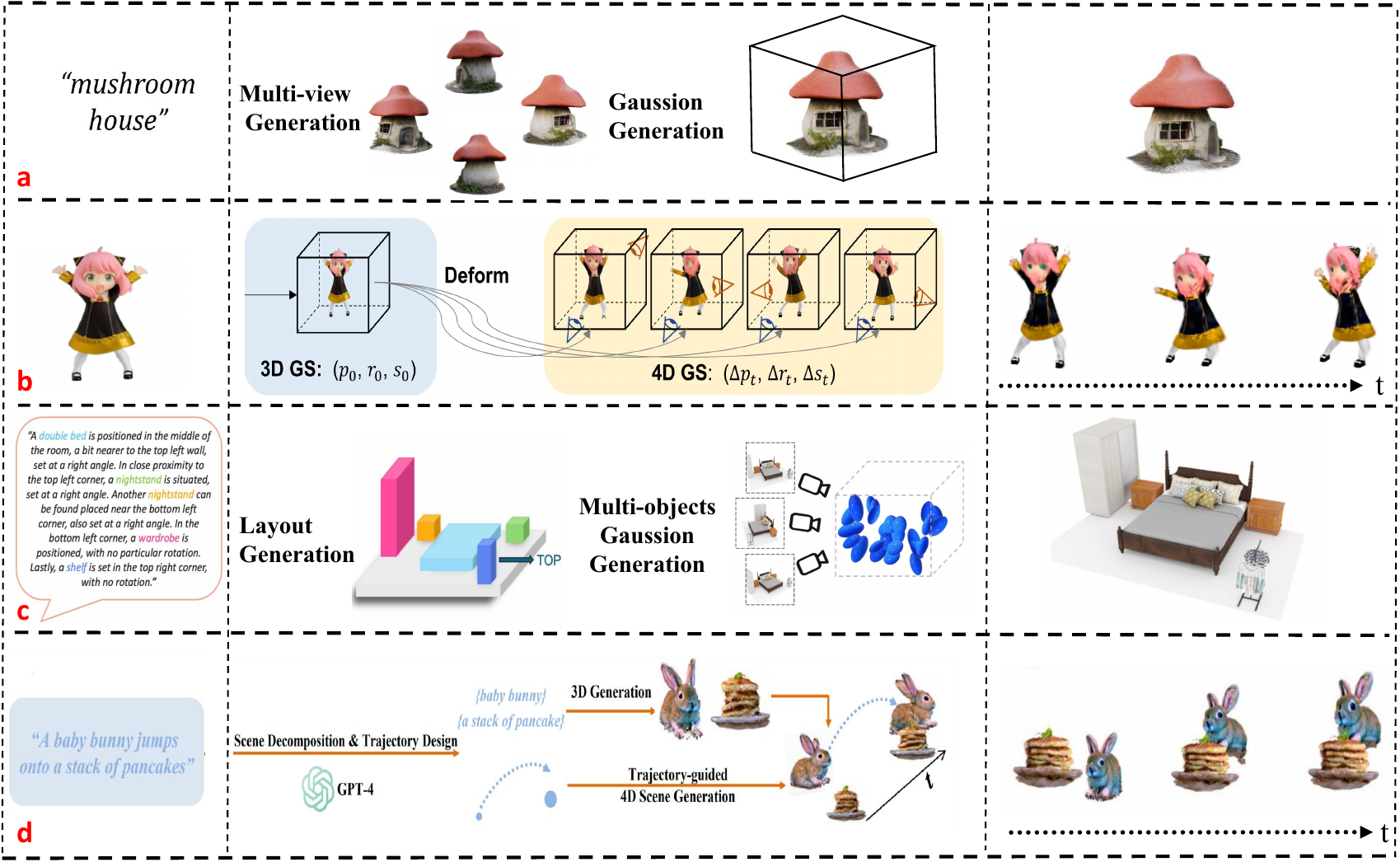}
    \caption{Illustration of NeRF and 3DGS applications in 3D content generation:
(a) 3D object-level generation~\cite{tang2024lgm}: a single 3D object is generated from a textual description;
(b) 4D object-level generation~\cite{ren2023dreamgaussian4d}: a dynamic 3D object (with temporal changes) is generated from an image input;
(c) 3D scene-level generation~\cite{ocal2024sceneteller}: a 3D indoor scene is synthesized from a textual description;
(d) 4D scene-level generation~\cite{xu2024comp4d}: a dynamic 3D scene is generated from a simple text prompt.}
    \label{fig:AIGC}
\end{figure*}

Thanks to their realistic rendering capabilities, Neural Radiance Fields (NeRF) and 3D Gaussian splatting (3DGS) have emerged as popular scene representations for 3D content generation. In contrast to many NeRF-based pipelines—which often require hours or even days of training—3DGS can reach comparable visual quality within minutes, or even seconds. Applications of NeRF and 3DGS can be broadly categorized into four areas: 3D object generation, 4D object generation, 3D scene generation, and 4D scene generation (Fig.~\ref{fig:AIGC}). The first column shows the input to the generation algorithm, the third column presents the generated 3D objects or scenes, and the second column illustrates typical generation pipelines. The different methods are compared in Table~\ref{tab:AIGC}.

\begin{table*}[htbp]
\centering
\begin{tabular}{|l|l|l|l|c|c|}
\hline
\textbf{Paper} & \textbf{Type} & \textbf{Input} & \textbf{Task} & \textbf{Year} & \textbf{FPS} \\
\hline
\multicolumn{6}{|c|}{\textbf{3D Object Generation}} \\
\hline
\cite{jain2022zero} & NeRF & Text & CLIP alignment; 3D object generation & CVPR 2022 & --- \\
\hline
\cite{li2024connecting} & GS & Text & Guided consistency sampling; 3D object generation & ECCV 2024 & --- \\
\hline
BrightDreamer~\cite{jiang2024brightdreamer} & GS & Text & Fast text-to-3D synthesis & arXiv 2024 & 77~ms \\
\hline
LGM~\cite{tang2024lgm} & GS & Text or image & Large multi-view Gaussian model & ECCV 2024 & --- \\
\hline
\cite{li2024controllable} & GS & Text and image & Controllable text-to-3D; SDS & 3DV 2025 & --- \\
\hline
Shape-E~\cite{jun2023shap} & NeRF & Text & Latent 3D Gaussian diffusion & arXiv 2023 & --- \\
\hline
L3DG~\cite{roessle2024l3dg} & GS & --- & Latent 3D Gaussian diffusion & SIGGRAPH Asia 2024 & $>10$~s \\
\hline
\multicolumn{6}{|c|}{\textbf{4D Object Generation}} \\
\hline
\cite{ling2024align} & GS & Text & Static 3D generation; deformation field & CVPR 2024 & --- \\
\hline
DreamGaussian4D~\cite{ren2023dreamgaussian4d} & GS & Image & 4D object generation; deformation network & arXiv 2023 & 5~min \\
\hline
Efficient4D~\cite{pan2024efficient4d} & GS & Video & Temporal and spatial image generation; 4D reconstruction & arXiv 2024 & 10~min \\
\hline
Consistent4D~\cite{jiang2024consistentd} & NeRF & Video & Video-to-4D; SDS & ICLR 2024 & --- \\
\hline
EG4D~\cite{sun2024eg4d} & GS & Image & Video generation; 4D generation & ICLR 2025 & --- \\
\hline
GaussianFlow~\cite{gao2024gaussianflow} & GS & Image & Image-to-video; Gaussian flow & TMLR 2025 & --- \\
\hline
\multicolumn{6}{|c|}{\textbf{3D Scene Generation}} \\
\hline
SceneTeller~\cite{ocal2024sceneteller} & GS & Text & 3D layout generation; 3D object generation & ECCV 2024 & --- \\
\hline
DreamScene~\cite{li2024dreamscene} & GS & Text & Formation pattern sampling; 3D scene generation & ECCV 2024 & $>10$~s \\
\hline
CG3D~\cite{vilesov2023cg3d} & GS & Text & Compositional 3D generation & arXiv 2023 & --- \\
\hline
HoloDreamer~\cite{zhou2024holodreamer} & GS & Text & Panorama generation; 3D reconstruction & arXiv 2024 & --- \\
\hline
Text2NeRF~\cite{zhang2024text2nerf} & NeRF & Text & Novel view synthesis; text-driven 3D generation & TVCG 2024 & --- \\
\hline
ART3D~\cite{li2024art3d} & GS & Text & Image generation; 3D reconstruction & CVPR 2024 & --- \\
\hline
RealmDreamer~\cite{shriram2024realmdreamer} & GS & Text & Text-to-image; 3D reconstruction & 3DV 2025 & --- \\
\hline
\cite{cai2024baking} & GS & Image & Diffusion models for image-to-3D & ICCV 2025 & --- \\
\hline
Flash3D~\cite{szymanowicz2024flash3d} & GS & Image & Depth prediction; 3D reconstruction & 3DV 2025 & --- \\
\hline
WonderWorld~\cite{yu2024wonderworld} & GS & Image & Novel view generation; 3D reconstruction & CVPR 2025 & --- \\
\hline
\multicolumn{6}{|c|}{\textbf{4D Scene Generation}} \\
\hline
4Real~\cite{yu20244real} & GS & Text & Canonical 3D representation; temporal deformation & NeurIPS 2025 & --- \\
\hline
Comp4D~\cite{xu2024comp4d} & GS & Text & Compositional 4D generation & arXiv 2024 & --- \\
\hline
DreamScene4D~\cite{chu2024dreamscene4d} & GS & Video & Decomposition; motion factorization; 4D composition & NeurIPS 2024 & --- \\
\hline
\end{tabular}
\caption{Comparison of NeRF-based and GS-based 3D content generation. ``FPS'' refers to throughput when explicitly reported; otherwise the entry shows per-sample latency or is not reported (---).}
\label{tab:AIGC}
\end{table*}

\subsection{3D Object Generation}
3D object generation methods can be primarily classified into three categories.

\textbf{(1) CLIP-guided optimization.} Dream Fields~\cite{jain2022zero} leverages CLIP’s cross-modal (text–image) alignment as semantic guidance: the NeRF representation is rendered from multiple virtual viewpoints to obtain 2D images, each rendering’s similarity to the input text is computed via CLIP, and the 3D parameters are optimized to maximize this score.

\textbf{(2) Score distillation sampling (SDS).} These approaches distill pretrained 2D text-to-image diffusion models (e.g., Imagen, Stable Diffusion) into 3D. \cite{li2024connecting} proposes consistency distillation and guided consistency sampling (GCS) to improve multiview consistency and semantic alignment. LGM~\cite{tang2024lgm} uses a feed-forward model that predicts high-resolution 3D Gaussians via cross-view fusion in a lightweight U-Net and trains end-to-end with differentiable rendering. \cite{li2024controllable} introduces a controllable framework with a multiview control network (MVControl) and a surface-aligned Gaussian representation (SuGaR) that binds Gaussians to mesh surfaces for improved geometry and texture. To enhance efficiency, BrightDreamer~\cite{jiang2024brightdreamer} presents a fully feed-forward pipeline that generates dense 3D Gaussian fields from text prompts using a text-guided shape-deformation network and a triplane generator, removing per-prompt optimization while maintaining fidelity and semantic alignment.

\textbf{(3) Direct diffusion over 3D primitives.} These methods avoid iterative SDS optimization by applying diffusion directly to 3D representations. Shape-E~\cite{jun2023shap} performs diffusion in the parameter space of implicit functions, decodable into textured meshes or NeRFs. L3DG~\cite{roessle2024l3dg} compresses 3D Gaussians into sparse, quantized latent features using sparse convolutions and vector quantization; at test time, novel objects are generated by denoising in the latent space.

\subsection{4D Object Generation}
4D object generation typically first produces a static 3D object (as above) and then learns a deformation network to animate motion~\cite{ling2024align,ren2023dreamgaussian4d,sun2024eg4d,gao2024gaussianflow}. DreamGaussian4D~\cite{ren2023dreamgaussian4d} generates multiview images from text to reconstruct a static 3D Gaussian model, then learns a temporal deformation field to animate Gaussians over time, guided by a text-driven video diffusion model, yielding high-quality 4D results within minutes. \cite{ling2024align} similarly trains a static 3D Gaussian from text and then learns a time-dependent MLP deformation field, optimized via differentiable rendering and SDS to maintain multiview, temporal consistency. EG4D~\cite{sun2024eg4d} starts from a single image, uses attention-injected video diffusion to produce temporally and multiview-consistent frames, fits a 4D Gaussian model with a color transformation network to decouple texture from geometry, and applies diffusion-based refinement—without SDS—to restore semantic details. GaussianFlow~\cite{gao2024gaussianflow} introduces “Gaussian flow,” mapping 3D Gaussian motion to pixel-level optical flow; through differentiable splatting, optical flow directly supervises spatiotemporal parameters, improving color stability and motion consistency. Efficient4D~\cite{pan2024efficient4d} uses a single-view video and an enhanced multiview diffusion model to synthesize consistent multiview sequences and trains a 4D Gaussian model capturing continuous motion. Consistent4D~\cite{jiang2024consistentd} employs a dynamic NeRF with a temporal consistency constraint for coherent video-to-4D reconstruction.

\subsection{3D Scene Generation}
3D scene generation follows two main paradigms. The first leverages large language models (LLMs) to generate structured scene layouts~\cite{ocal2024sceneteller,li2024dreamscene,vilesov2023cg3d}, which are then instantiated and rendered with 3D Gaussians. The second bypasses explicit layouts by directly synthesizing multi-object RGB and depth images from text~\cite{zhou2024holodreamer,li2024art3d,shriram2024realmdreamer,ouyang2023text2immersion,cai2024baking,szymanowicz2024flash3d,yu2024wonderworld} and reconstructing the scene using a 3DGS pipeline.

SceneTeller~\cite{ocal2024sceneteller} uses an LLM to propose a structured layout from text, retrieves assets from a 3D database, renders multiview images, and reconstructs with Gaussian splatting; text-driven stylization can be applied to the resulting Gaussians. DreamScene~\cite{li2024dreamscene} parses object semantics, spatial relations, and environment descriptors; formation pattern sampling sets object layouts and attributes, each object is optimized as a 3D Gaussian, and simple primitives (e.g., cuboids, hemispheres) compose the environment before a final refinement stage. CG3D~\cite{vilesov2023cg3d} converts language into a scene graph and instantiates each object via Gaussian splatting for modular, controllable composition.

HoloDreamer~\cite{zhou2024holodreamer} forgoes explicit layouts by generating stylized panoramas via multiple diffusion models, estimating monocular depth, and running two-stage Gaussian optimization; because panoramas often include multiple objects and humans, rich scenes are obtained without retrieval. Text2NeRF~\cite{zhang2024text2nerf} follows a similar pipeline with a NeRF representation. ART3D~\cite{li2024art3d} improves depth estimation for better structural consistency. DreamScene360~\cite{zhou2024dreamscene360} targets immersive \(360^\circ\) generation using quality assessment and prompt revision to select candidates and panoramic Gaussian splatting with semantic–geometric guidance for reconstruction. RealmDreamer~\cite{shriram2024realmdreamer} initializes scenes from synthesized RGB and depth and enhances them via inpainting and depth distillation. Text2Immersion~\cite{ouyang2023text2immersion} generates an initial 3D scene from synthesized RGB/depth, then samples extra viewpoints and refines via diffusion-based inpainting. \cite{cai2024baking} integrates Gaussian splatting into a diffusion denoiser to produce 3D Gaussian point clouds end-to-end, enforcing angular constraints on viewpoint vectors. Flash3D~\cite{szymanowicz2024flash3d} predicts metric depth and Gaussian parameters with an encoder–decoder network and uses depth offsets to refine centers. WonderWorld~\cite{yu2024wonderworld} reconstructs a 3D scene from a single image and supports interactive text-driven expansion via diffusion-based view extension.

\subsection{4D Scene Generation}
4D scene methods typically first construct a canonical 3D Gaussian representation and then drive object motion~\cite{yu20244real,xu2024comp4d}. 4Real~\cite{yu20244real} uses a text-to-video diffusion model to create a reference video with dynamics, generates a freeze-time video (circular camera, minimal motion), reconstructs a canonical 3D representation, and finally learns temporal deformations to match the reference motion. Comp4D~\cite{xu2024comp4d} uses an LLM for scene decomposition into individual 3D objects and for trajectory design, guiding compositional 4D optimization; the final 4D scene is rendered with 3D Gaussians. DreamScene4D~\cite{chu2024dreamscene4d} decomposes and amodally completes each object and the background, fits static 3D Gaussians, factorizes and optimizes per-object motion, and recomposes independently optimized 4D Gaussians into a unified frame using estimated monocular depth.

\begin{figure*}[h]
  \centering
  \includegraphics[width=\textwidth]{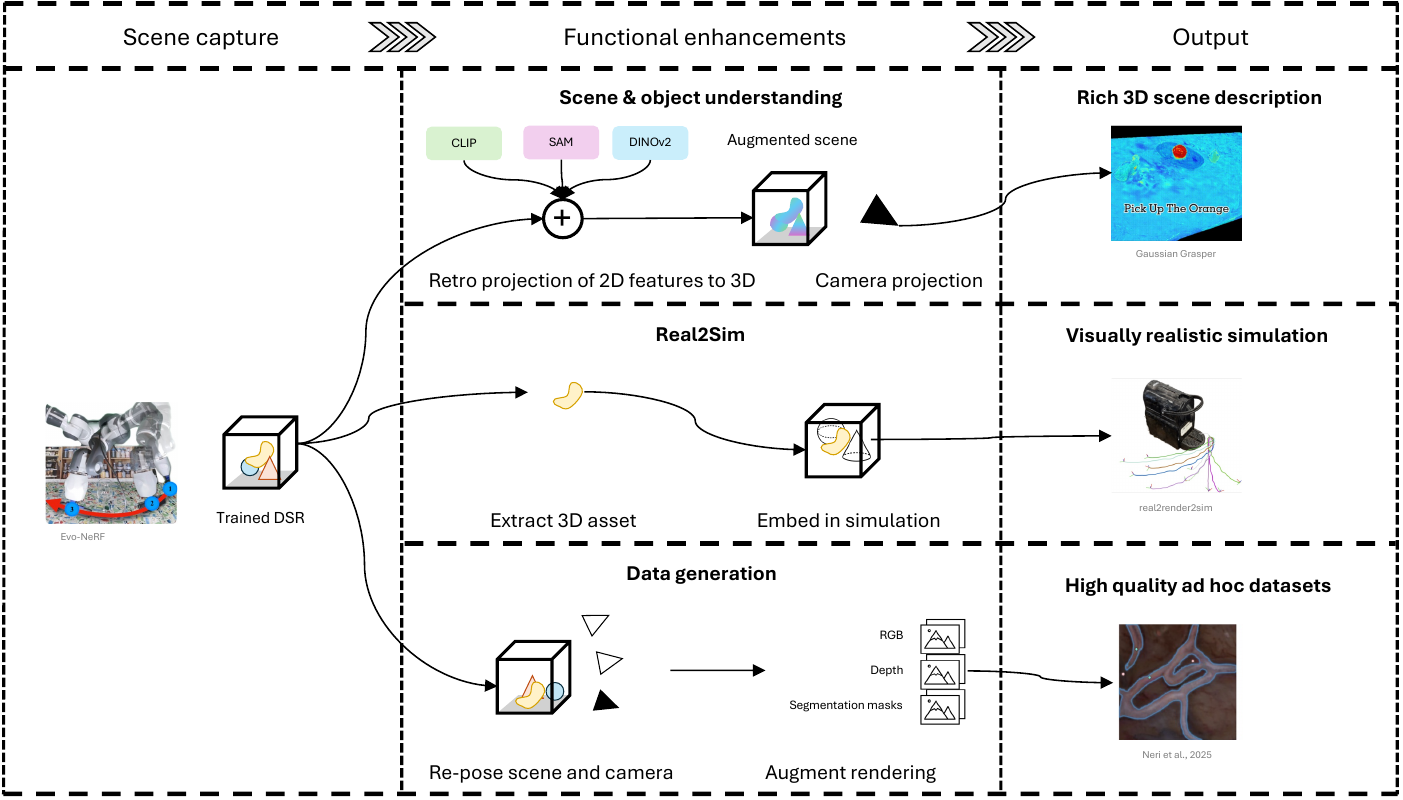}
  \caption{Differentiable scene representations such as radiance fields and Gaussian splats have a wide array of applications for robotic manipulation. The trained scene representation can support rich 3D scene descriptions~\cite{zheng2024gaussiangrasper3dlanguagegaussian}, serve as realistic assets for simulation training~\cite{yu2025real2render2realscalingrobotdata}, or act as synthetic data generators~\cite{neri2025surgical}.}
  \label{fig:robo_manip}
\end{figure*}

\section{Robotic Manipulation}
\label{sec:manipulation}
Autonomous robotic manipulation hinges on a robot’s ability to infer actionable geometric and semantic information from its environment. Traditional pipelines rely on modular perception, where pretrained components extract low-dimensional intermediates—such as object masks, 6D poses, or affordance scores—from RGB-D or point-cloud data. Although effective, these pipelines often incur information bottlenecks, limited spatial continuity, and brittle integration with downstream planning.

Differentiable scene representations (DSRs) propose an alternative: represent the environment as a continuous, gradient-compatible structure that can be optimized end-to-end. Unlike modular pipelines, DSRs embed visual, geometric, and semantic information in a unified model, enabling joint optimization across perception, memory, and control. Because the rendering process is differentiable, DSRs naturally handle occlusion and exploit multiview consistency, which improves generalization across viewpoints and over time.

In this section, we outline how recent DSR methods impact robotic manipulation. An overview is shown in Fig.~\ref{fig:robo_manip}, and representative methods are compared in Table~\ref{tab:robotic_manipulation}.

\begin{table*}[!t]
\centering
\begin{tabular}{|l|l|l|l|c|c|}
\hline
\textbf{Paper} & \textbf{Type} & \textbf{Input} & \textbf{Task} & \textbf{Year} & \textbf{FPS} \\
\hline
\multicolumn{6}{|c|}{\textbf{Object-level modeling}}\\
\hline
Dex-NeRF~\cite{IchnowskiAvigal2021DexNeRF} & NeRF & RGB & \multirow{3}{*}{Depth map rendering; grasp planning} & CoRL 2021 & --- \\
Evo-NeRF~\cite{kerr2022evo} & NeRF & RGB &  & CoRL 2022 & $<1$ \\
Transplat~\cite{kim2025transplat} & GS & RGB-D &  & arXiv 2025 & --- \\
\cite{han2025nerfbasedtransparentobjectgrasping} & NeRF; SDF & Image & Pose estimation; grasp prediction & arXiv 2025 & --- \\
POGS~\cite{yu2025persistent} & GS & RGB-D & Object tracking; manipulation; scene editing & ICRA 2025 & --- \\
\hline
\multicolumn{6}{|c|}{\textbf{Scene understanding}}\\
\hline
\cite{li2024object} & GS & RGB & Language-conditioned dynamic grasping & \makecell{ICRA 2024\\Workshop} & 30 \\
Gaussian Grasper~\cite{zheng2024gaussiangrasper3dlanguagegaussian} & GS & RGB-D & Grasp planning; natural-language instructions & arXiv 2024 & 5.3 \\
Splat-MOVER~\cite{shorinwa2024splat} & GS & RGB (monocular) & Grasp planning; scene editing; natural-language instructions & CoRL 2024 & --- \\
ManiGaussian~\cite{lu2024manigaussian} & GS & RGB-D & Scene dynamics prediction; natural-language instructions & ECCV 2024 & --- \\
GraspSplats~\cite{ji2024-graspsplats} & GS & RGB-D & Grasp sampling; scene editing; natural-language instructions & CoRL 2024 & --- \\
\hline
\multicolumn{6}{|c|}{\textbf{Data generation}}\\
\hline
Real2Render2Real~\cite{yu2025real2render2realscalingrobotdata} & GS & RGB & \multirow{2}{*}{\makecell[l]{Data generation; imitation learning; data augmentation;\\ real2sim2real}} & arXiv 2025 & --- \\
$Re^3$Sim~\cite{han2025re3sim} & GS & RGB-D &  & arXiv 2025 & --- \\
GARField~\cite{delehelle2024garfield} & NeRF & RGB-D & \multirow{2}{*}{Data generation; real2sim} & ROBIO 2024 & $<1$ \\
COV-NeRF~\cite{mishra2024closing} & NeRF & RGB-D &  & ICRA 2024 & $<1$ \\
\cite{neri2025surgical} & NeRF & RGB & Single-image training; surgical data generation & IJCARS 2025 & $<1$ \\
\hline
\end{tabular}
\caption{Comparison of differentiable rendering methods for robotic manipulation. ``FPS'' reports throughput when explicitly given; otherwise the entry is not reported (---).}
\label{tab:robotic_manipulation}
\end{table*}

\subsection{Traditional vs.\ Differentiable Scene Representations}
Traditional perception in autonomous manipulation typically comprises a stack of modules (e.g., object detectors~\cite{xiang2017posecnn}, semantic segmenters~\cite{gao2021kpam}, or keypoint extractors~\cite{tian2025yolov12attentioncentricrealtimeobject}) trained on labeled datasets and applied independently of the robot’s planner or controller. These systems often operate in image space or projective 3D space, yielding discrete and brittle outputs such as bounding boxes or class masks. Additionally, outputs are independent, so modules do not interact to produce consolidated estimates.

In contrast, DSRs such as Neural Radiance Fields (NeRFs), signed distance fields (SDFs), or Gaussian splatting represent scenes volumetrically and make the rendering process explicit upstream of pixel supervision. This enables tight coupling between perception and downstream components, facilitating gradient flow from task losses to scene parameters. DSRs also provide multiview aggregation, occlusion-aware reconstruction, and joint reasoning over geometry and semantics. Finally, geometric regularization can be added to stabilize training and improve consistency.

Nevertheless, DSRs are not without drawbacks. High computational costs during training, limited real-time performance, and data inefficiency remain open challenges. Selecting the appropriate representation entails tradeoffs among fidelity, latency, and memory footprint.

\subsection{Object-Level Modeling and Scene Understanding}
Perception for \emph{transparent} objects illustrates the benefit of tightly coupling geometry and appearance. Transparent items exhibit weak texture, view-dependent specularities, and refractive effects; depth sensing is often noisy and inconsistent across frames. Dex-NeRF~\cite{IchnowskiAvigal2021DexNeRF} builds a dataset with controlled lighting to enhance specular cues, trains a NeRF to reconstruct the scene, and renders depth maps to support grasp planning. Evo-NeRF~\cite{kerr2022evo} adds geometric regularization and improves grasp prediction to better exploit NeRF-derived depth. \textbf{Transplat}~\cite{kim2025transplat} combines a latent diffusion model for dense surface embeddings with Gaussian splatting for cohesive 3D reconstruction, reducing depth error across two synthetic and one real dataset versus four baselines. \cite{han2025nerfbasedtransparentobjectgrasping} further improves NeRF-based depth by injecting shape priors at capture.

Beyond transparency, DSRs benefit general grasping and manipulation. POGS~\cite{yu2025persistent} distills 2D foundation-model outputs into an object-level 3D Gaussian map for robust tracking, grasping, reorientation, and natural language–conditioned manipulation. For \emph{scene}-level understanding, works such as Gaussian Grasper~\cite{zheng2024gaussiangrasper3dlanguagegaussian}, Splat-MOVER~\cite{shorinwa2024splat}, GraspSplats~\cite{ji2024-graspsplats}, and \cite{li2024object} consolidate semantics from CLIP~\cite{radford2021learning} and masks from SAM~\cite{kirillov2023segment} into a common 3D Gaussian scene, providing a rich substrate for fast, high-level decision making. Dynamics can also be modeled: ManiGaussian~\cite{lu2024manigaussian} projects Perceiver IO~\cite{jaegle2021perceiver} outputs into a dynamic Gaussian map, giving the planner foresight about action outcomes.

\subsection{Motion Planning With Differentiable Scene Representations}
For motion planning, signed/unsigned distance fields are popular because they provide smooth, dense distance gradients throughout the workspace. ReDSDF~\cite{9981456} represents scene objects and the human body with SDFs for reactive motion generation. DCPF~\cite{10670290} addresses the cost of sampling implicit scene models by learning deep collision probability fields, shifting compute to training and enabling faster inference. Note that many recent differentiable approaches operate in the robot’s \emph{configuration space}~\cite{10611674,chi2024safe,9976191,10238810,10610773,long2025neuralconfigurationspacebarriersmanipulation}—outside the scope here, as they do not directly provide a scene representation.

\subsection{Data Generation}
Another promising use of DSRs is upstream of planning, for realistic data generation. Policies trained in simulation often struggle to transfer due to visual and physical sim–real gaps. Because DSRs learn directly from real imagery and support photorealistic novel-view synthesis or controlled scene editing, they can generate tailored datasets for downstream learning. $Re^3$Sim~\cite{han2025re3sim} and Real2Render2Real~\cite{yu2025real2render2realscalingrobotdata} use Gaussian splatting for fast data generation to improve transfer in rigid-object manipulation. COV-NeRF~\cite{mishra2024closing} adopts a NeRF-based pipeline for cluttered manipulation. NeRFs are also used where labeled data are scarce and costly, e.g., surgical robotics~\cite{neri2025surgical} and garment manipulation in GARField~\cite{delehelle2024garfield}.

\section{Discussion and Future Opportunities}
\label{sec:challenges}
Despite rapid progress in neural scene representations, significant challenges remain in NeRF- and 3DGS-based methods. In this section, we summarize common technical challenges observed across domains and highlight promising research opportunities.

\subsection{Latency and Real-Time Constraints}
NeRF-based methods remain computationally intensive due to volumetric ray marching, limiting their use in time-critical scenarios such as SLAM and telepresence. Even with hash encoding or distillation, incremental updates are slow and memory demanding. By contrast, 3DGS achieves interactive frame rates through rasterization of Gaussian primitives, but scaling to larger environments or high-resolution outputs increases memory overhead and training time.

Future work should explore hybrid pipelines in which compact NeRF models serve as global priors while 3DGS handles online rendering and updates. Adaptive splat compression, foveated rendering, and level-of-detail (LOD) control will be critical for bandwidth-limited scenarios such as teleoperation and edge robotics.

\subsection{Geometry and Scene Representation}
NeRF excels at view-dependent appearance but struggles with precise geometry extraction, which is essential for planning and robotic manipulation. 3DGS provides explicit primitives that are easy to rasterize but can yield discontinuous or noisy reconstructions. Neither representation currently achieves robust, mesh-quality geometry suitable for safety-critical control.

Promising directions include combining 3DGS with signed distance fields (SDFs) or surface regularizers for mesh extraction, developing uncertainty-aware geometry estimation, and devising representations that maintain both photorealism and planning-safe geometry. Such advances are particularly relevant for autonomous manipulation and SLAM back ends.

\subsection{Dynamic and Deformable Environments}
Both NeRF and 3DGS methods are primarily designed for static scenes. Real-world robotics and telepresence involve dynamic humans, moving objects, and changing illumination. NeRF variants with deformation fields incur high latency, while 3DGS-based methods face difficulty maintaining temporal consistency and avoiding primitive explosion in unobserved regions.

Researchers could extend 3DGS with temporally coherent primitives (e.g., velocity or deformation codes) and integrate event sensing or inertial measurement units (IMUs) to enable real-time adaptation in dynamic settings. These improvements are essential for human–robot collaboration, immersive telepresence, and interactive 3D content generation.

\subsection{Scalability and Efficiency}
NeRF compresses appearance into compact networks but scales poorly in large or long-term scenarios due to retraining costs. 3DGS enables real-time rendering but may require millions of primitives, leading to high memory usage and storage demands, especially for city-scale SLAM or long-duration teleoperation.

Promising directions include hierarchical Gaussian structures, streaming representations that transmit only relevant subsets, and rate–distortion–optimized codecs for Gaussian primitives. These would enable efficient deployment in cloud–edge robotics pipelines and XR systems.

\subsection{Uncertainty and Robustness}
Neither NeRF nor 3DGS inherently models uncertainty. This hinders their use in safety-critical manipulation or navigation, where reliable confidence estimates are required. Robustness to sensor noise, occlusion, and poor initialization also remains limited, especially in unconstrained telepresence or outdoor SLAM.

Future work should integrate probabilistic formulations into 3DGS to enable confidence-aware updates and risk-sensitive control. Coupling these models with semantic priors or large foundation models may further enhance robustness in complex environments.

\subsection{Integration with Robotics and Multimodal Systems}
Current pipelines are primarily vision-driven, with limited integration of LiDAR, tactile sensing, or semantics. This restricts deployment in tasks where multimodal information is essential, such as manipulation under occlusion or teleoperation with limited visibility.

Incorporating multimodal cues into NeRF and 3DGS pipelines could yield richer, more reliable scene representations. Combining 3DGS with foundation models for language and vision can further enable task-level reasoning, scene annotation, and instruction-driven control—key enablers for cross-domain deployment.

\subsection{Standardization and Benchmarks}
Evaluation across studies remains inconsistent, with most focusing on visual fidelity (PSNR, SSIM) rather than task-specific metrics such as localization drift, control accuracy, or latency. Establishing standardized, latency-centric benchmarks that cover SLAM, telepresence, 3D content generation, and robotic manipulation is essential. Such benchmarks should evaluate not only rendering quality but also responsiveness, robustness, and downstream task performance.

Overall, while NeRF provides compact global appearance modeling, 3DGS has emerged as a practical candidate for real-time deployment due to its explicit representation and fast rasterization. The future lies in hybrid pipelines that exploit the strengths of both, combined with advances in dynamic scene modeling, compression, multimodal integration, and benchmark-driven evaluation. These directions will help move neural scene representations from academic novelty to reliable components of robotics and immersive systems.

\section{Conclusion}
\label{sec:conclusions}
Neural scene representations have evolved rapidly, moving from optimization-heavy volumetric fields to fast, explicit, and differentiable structures. This survey analyzed the impact of Neural Radiance Fields (NeRF) and 3D Gaussian Splatting (3DGS) relative to prior explicit approaches and feed-forward predictors, with an emphasis on integration across SLAM, telepresence, 3D content generation, and robotic manipulation.

Across these domains, 3DGS offers a strong balance among photorealism, rendering speed, and structural flexibility. It provides a practical foundation for tasks that require both visual quality and downstream usability, including dense mapping, real-time rendering, sim-to-real transfer, and editable content generation. We highlighted how 3DGS is increasingly replacing NeRF-based systems in interactive pipelines and how it enables hybrid architectures that combine geometry, learning, and semantics.

Important challenges remain: improving temporal consistency in dynamic scenes, reducing memory overhead at scale, and integrating semantic priors for task-aware reasoning. Designing hybrid pipelines that combine 3DGS with transformers, diffusion models, or classical geometry is a promising direction. We hope this survey helps researchers and practitioners navigate a fast-moving field and provides a roadmap for building high-performance neural 3D systems across embodied and generative applications.

\section{Financial Support}
This research is supported by and in collaboration with the Italian National Institute for Insurance against Accidents at Work (INAIL), under the project ``ROBOT INSPECTION ASSISTITA DI ATTREZZATURE A PRESSIONE''.

{\small
\bibliographystyle{unsrt}
\bibliography{references}
}
\vspace{-1cm}

\begin{IEEEbiography}
[{\includegraphics[width=1in,height=1.25in,clip,keepaspectratio]{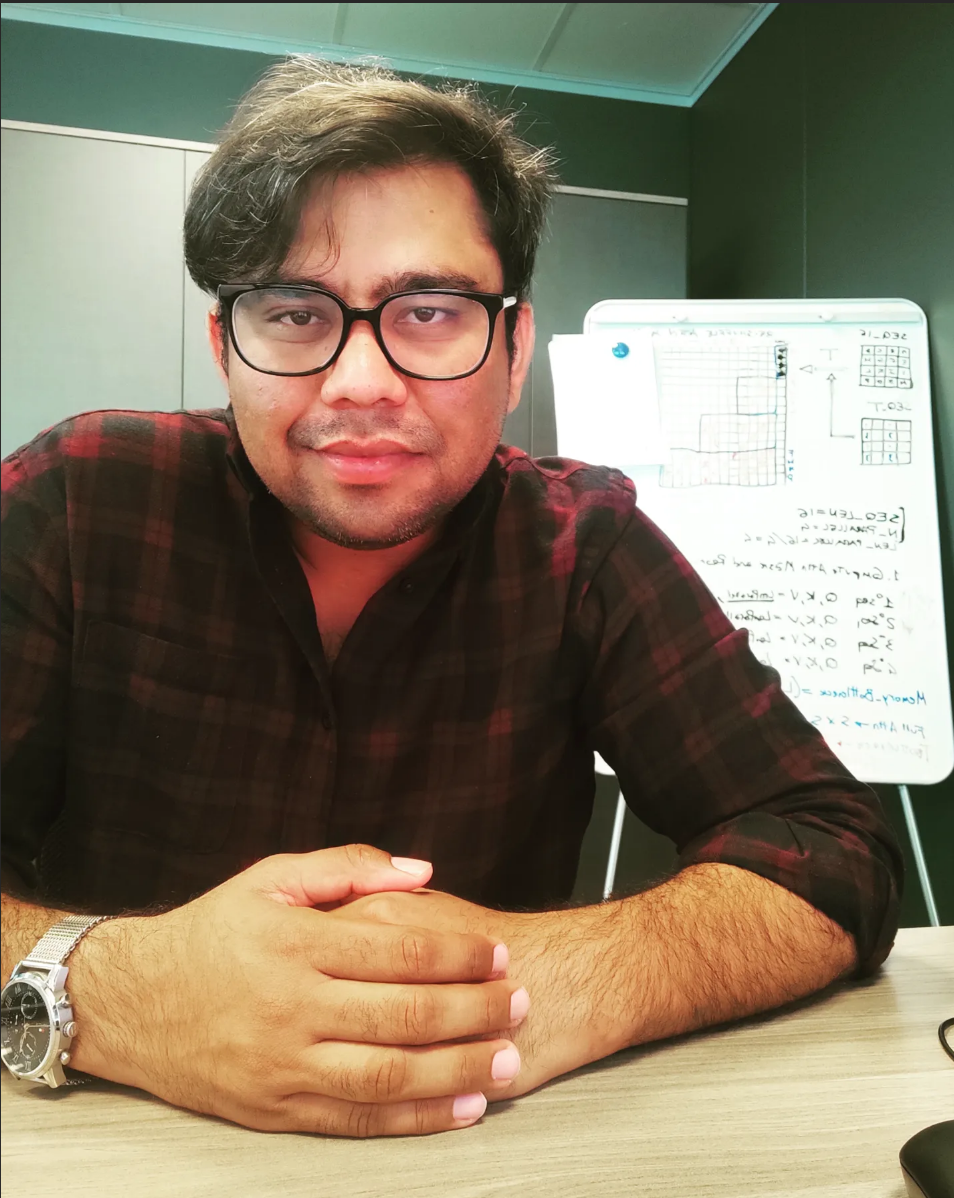}}]{\textbf{Javed Ahmad}}
received the Ph.D. degree in computational vision, automatic recognition, and learning from the University of Genoa, Italy, in 2023. He is currently a Postdoctoral Researcher with the Advanced Robotics Department, Istituto Italiano di Tecnologia (IIT), Italy. His research interests include 3D computer vision, multimodal scene understanding, SLAM, and robotics, with a focus on low-latency, real-time systems. He has served as a reviewer for IEEE Robotics and Automation Letters, ICRA, and IROS.
\end{IEEEbiography}
\vspace{-1cm}
\begin{IEEEbiography}
[{\includegraphics[width=1in,height=1.25in,clip,keepaspectratio]{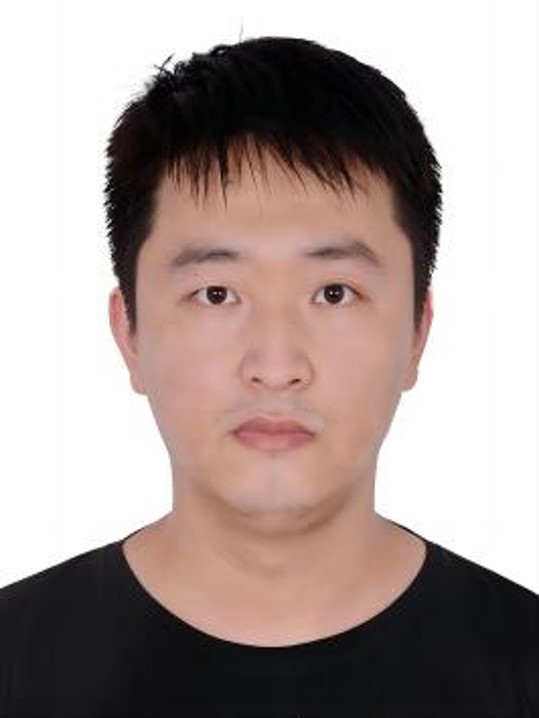}}]{\textbf{Penggang Gao}}
is currently pursuing the Ph.D. degree with the Istituto Italiano di Tecnologia (IIT) and the University of Genoa, Italy. He previously worked as a research assistant at the University of Hong Kong. His research interests include 3D scene understanding, 3D scene generation, and their applications in robotics.
\end{IEEEbiography}
\vspace{-1cm}
\begin{IEEEbiography}
[{\includegraphics[width=1in,height=1.25in,clip,keepaspectratio]{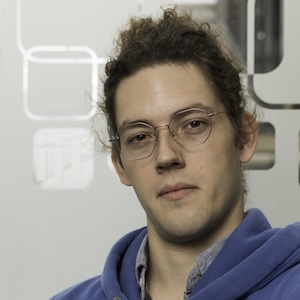}}]{\textbf{Donatien Delehelle}}
is a Ph.D. student at IIT and the University of Genoa, Italy. He received the master's degree in data science from \'{E}cole Centrale de Lyon, France, in 2020. He is currently working on garment manipulation under the supervision of Prof. Darwin G. Caldwell (IIT) and Dr. Fei Chen (The Chinese University of Hong Kong). His research interests include deep learning, self-supervised learning, reinforcement learning, and deformable object manipulation.
\end{IEEEbiography}
\vspace{-1cm}
\begin{IEEEbiography}
[{\includegraphics[width=1in,height=1.25in,clip,keepaspectratio]{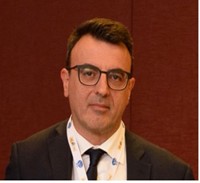}}]{\textbf{Mennuti Canio}}
holds a civil engineering degree from the University of Naples “Federico II.” Since 1994, he has been involved in assessing and verifying the safety requirements of pressure equipment and, more generally, industrial plants at the Higher Institute for Prevention and Safety at Work (ISPESL). Since 2002, he has been engaged in research at the National Institute for Insurance against Accidents at Work (INAIL) in the field of non-destructive testing and plant integrity. Since 2015, he has headed the Innovative Technologies for Safety Laboratory of the INAIL Technological Innovations Department. He is the author of numerous scientific publications and co-inventor of an industrial patent.
\end{IEEEbiography}
\vspace{-1cm}
\begin{IEEEbiography}
[{\includegraphics[width=1in,height=1.25in,clip,keepaspectratio]{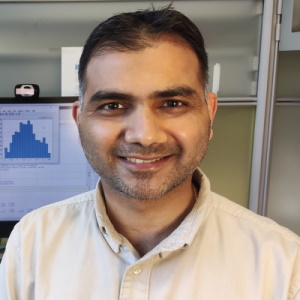}}]{\textbf{Nikhil Deshpande}} (Member, IEEE)
received the Ph.D. degree in electrical engineering from North Carolina State University, Raleigh, NC, USA, in 2012. He is currently an Associate Professor of Robotics and AI with the Cyber-physical Health and Assistive Robotics Technologies (CHART) Group, University of Nottingham, U.K. He has more than 15 years of experience in robotics and artificial intelligence, with research interests spanning surgical robotics, telerobotics, mixed reality, machine learning, and computer vision. Prior to joining Nottingham, he was a Researcher at the Istituto Italiano di Tecnologia (IIT), where he led the VICARIOS Mixed Reality and Simulations Lab. He has served as an editor and reviewer for major IEEE conferences and journals, including ICRA, IROS, T-RO, and RAS.
\end{IEEEbiography}
\vspace{-1cm}
\begin{IEEEbiography}
[{\includegraphics[width=1in,height=1.25in,clip,keepaspectratio]{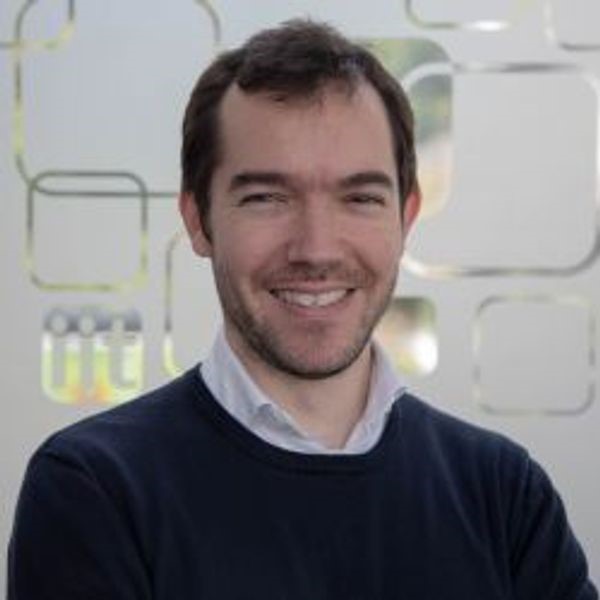}}]{\textbf{Jes\'us Ortiz}}
received the M.Sc. degree in mechanical engineering from the University of Zaragoza, Spain, in 2003, and the Ph.D. degree in transport systems and vehicles from the University of Zaragoza in 2008. Since 2006, he has been with the Istituto Italiano di Tecnologia (IIT), Genoa, Italy, where he currently heads the Wearable Robots, Exoskeletons and Exosuits Laboratory (XoLab), Department of Advanced Robotics (ADVR). His research interests include motion bases, driving simulators, teleoperation, telepresence, virtual reality, general-purpose GPU computing, medical robotics, wearable robotics, industrial exoskeletons, and soft wearable assistive devices.
\end{IEEEbiography}
\vspace{-1cm}
\begin{IEEEbiography}
[{\includegraphics[width=1in,height=1.25in,clip,keepaspectratio]{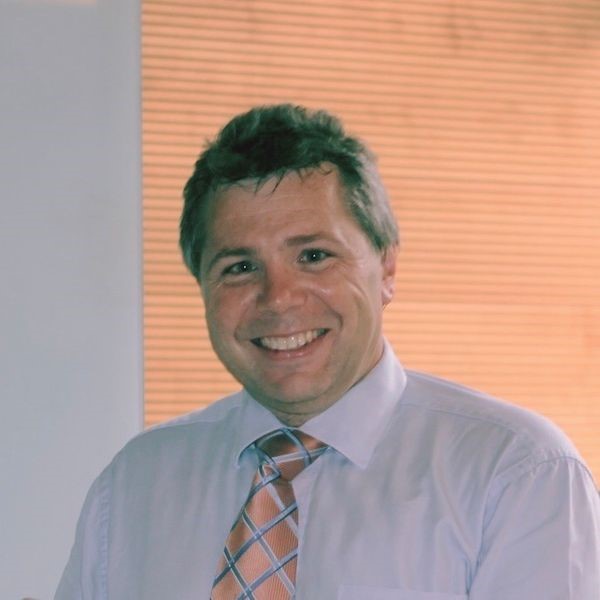}}]{\textbf{Darwin G. Caldwell}} (Member, IEEE)
received the B.Sc. and Ph.D. degrees in robotics from the University of Hull, Hull, U.K., in 1986 and 1990, respectively, and the M.Sc. degree in management from the University of Salford, Salford, U.K., in 1994. He is a Deputy Director with the Istituto Italiano di Tecnologia (IIT), Genoa, Italy, where he is also the Founding Director of the Department of Advanced Robotics. He is or has been an Honorary Professor with the Universities of Manchester, Sheffield, and Bangor, U.K.; King’s College London, U.K.; Tianjin University, China; and Shenzhen Academy of Aerospace Technology, China. He is the author or coauthor of more than 500 academic papers and more than 20 patents. His research interests include medical, rehabilitation, and assistive robotic technologies; humanoids; innovative actuators; and haptics and force-augmentation exoskeletons. Prof. Caldwell was the recipient of 40 awards and nominations from leading journals and conferences.
\end{IEEEbiography}
\vspace{-0.5cm}
\begin{IEEEbiography}
[{\includegraphics[width=1in,height=1.25in,clip,keepaspectratio]{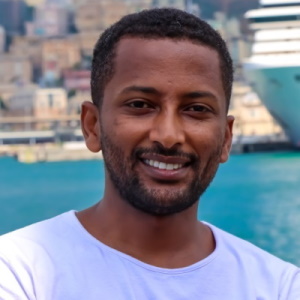}}]{\textbf{Yonas Teodros Tefera}}
holds the B.Sc. degree in electrical and computer engineering, the M.Sc. degree in information and communication engineering from the University of Trento, Italy, and the Ph.D. degree from the Istituto Italiano di Tecnologia (IIT) and the University of Verona (UNIVR), Italy. He is now the research team lead of the VICARIOS Lab in the Advanced Robotics research line at IIT. His research focuses on developing immersive teleoperation and telepresence techniques to improve remote presence and situational awareness, using robotics, haptics, 3D perception, gaze-driven rendering, and advanced 2D/3D compression and streaming.
\end{IEEEbiography}


\end{document}